# Tailored robotic training improves hand function and proprioceptive processing in stroke survivors with proprioceptive deficits: A randomized controlled trial

**Authors:** Andria J. Farrens[1], Luis Garcia-Fernandez[2], Raymond Diaz Rojas[2], Jillian Obeso Estrada[3], Dylan Reinsdorf[2], Vicky Chan, DPT[4], Disha Gupta[5], Joel Perry[6], Eric Wolbrecht[6], An Do, MD[7], Steven C. Cramer, MD[8], David J. Reinkensmeyer[2,3,9].

**Affiliations:**

1: Department of Orthopedics and Rehabilitation, Division of Physical Therapy; University of New Mexico, Albuquerque

2: Department of Mechanical and Aerospace Engineering, UC Irvine, Irvine, CA

3: Department of Biomedical Engineering, UC Irvine, Irvine, CA

4: Department of Physical Therapy, UC Irvine, Irvine, CA

5: National Center for Adaptive Neurotechnologies, Stratton VA Medical Center; Electrical and Computer Engineering Dept. Albany State University of New York

6: Department of Mechanical Engineering, University of Idaho, Moscow, USA

7: Department of Neurology, UC Irvine; Irvine, CA

8: Department of Neurology, UCLA; California Rehabilitation Institute; Los Angeles, CA

9: Department of Anatomy and Neurobiology, UC Irvine, Irvine, CA

## ABSTRACT

**Background**: Precision rehabilitation aims to tailor movement training to improve outcomes. Robots provide a platform to test tailored approaches. Standard robotic training methods provide physical assistance during visually guided, game-based tasks. This approach is ineffective for individuals with impaired proprioception (~50% of stroke survivors). Here, we evaluated two novel proprioceptively-tailored approaches: Propriopixel Training, which uses robot-facilitated movements as gamified-training cues to enhance proprioceptive perception during motor planning, and Virtual Assistance Training, which removes physical assistance to increase reliance on self-generated proprioceptive feedback. **Methods**: 46 individuals with chronic stroke and hand impairment were randomized to Standard, Propriopixel, or Virtual Assistance training using the FINGER robot. Participants completed nine two-hour sessions (~1050 movements/session at ~80% success) over three weeks. Box and Blocks Test (BBT) was the primary outcome; secondary measures included robotic measures of proprioception monitored with EEG. **Results**: All groups (baseline BBT=24 [3-53] blocks) improved comparably in BBT (4 ± 6.7 blocks). For individuals with proprioceptive deficits, Propriopixel (7 ± 4.2 blocks, p=0.002) and Virtual Training (4.5 ± 4.4 blocks, p=0.069) outperformed Standard Training (0.8 ± 2.3 blocks). Only Propriopixel and Virtual Training yielded proprioceptive gains, which correlated with BBT improvements. A novel EEG measure, the proprioceptive Contingent Negative Variation, indicated these approaches enhanced neural sensitivity to proprioceptive cues, particularly estimation of finger velocity. **Conclusion**: Training response depended on initial proprioceptive status, and novel proprioceptively-tailored robotic training significantly improved hand function and proprioceptive processing in individuals with proprioceptive deficits, supporting a new approach to precision movement rehabilitation after stroke.

**One sentence summary:** Proprioceptive-focused robotic training improved hand function and sensation in chronic stroke survivors, advancing precision movement rehabilitation.

## INTRODUCTION

Advancing stroke rehabilitation requires a shift from standardized treatment to personalized approaches tailored to individual differences in recovery(*1*). Robotic therapy devices provide a means to evaluate novel personalized interventions with a high degree of control. While numerous robotic training strategies have been developed for the stroke-affected upper extremity(*2–4*) (UE) active "assist-as-needed" physical assistance is the most commonly researched and commercially deployed approach(*5–8*). In active assistance paradigms, the patient must initiate movement to trigger physical assistance, pairing movement intent with subsequent action to stimulate motor learning circuits(*1, 2*). Active assistance necessitates patient engagement and effort, improves motivation to participate in training(*9, 10*), and expands movement range of motion and accuracy, enhancing proprioceptive input(*11–13*).

Systematic reviews indicate that this standard robotic training approach modestly reduces UE impairment, with outcomes comparable to or slightly better than dose-matched conventional therapy(*3, 14, 15*). The degree of benefit varies substantially between individuals, and the underlying mechanisms contributing to this response variability are poorly understood, hindering a precision rehabilitation approach(*16–18*).

Recently, we found that proprioception—the sense of position and movement—is a key factor related to clinical outcomes in robotic hand therapy(*19, 20*). Both impaired finger proprioception (measured with a novel robotic paradigm) and injury to the somatosensory system were associated with reduced UE therapeutic gains following standard robotic training. This adds to a growing body of evidence, including the EXCITE trial of constraint-induced therapy, that individuals with impaired UE proprioception benefit significantly less from existing forms of UE movement therapy(*21–24*).

We hypothesize that proprioception contributes to motor rehabilitation through Hebbian-based learning mechanisms that reinforce neural pathways through repeated co-activation of sensory and motor signals(*25*). Theoretical models and experimental evidence suggest that the sensory input experienced during movement practice paired with voluntary attempts to activate motor pathways can drive sensorimotor reorganization(*26–29*). In theory, physical assistance helps stimulate this type of learning by allowing participants to perform and receive sensory feedback of improved movement patterns(*30, 31*). However, when proprioceptive processing is impaired, such feedback may not be appropriately attended to and interpreted by the sensorimotor system, limiting learning. Proprioception is estimated to be impaired in over half of chronic stroke survivors(*22, 32*). Given the prevalence and potential role of proprioceptive deficits in responsiveness to standard robotic training, we aimed to develop new training paradigms for individuals with proprioceptive deficits.

In the first robotic training paradigm focused on proprioceptive training, we developed a Propriopixel training mode that requires increased attention to proprioceptive feedback during physical assistance-based training. In standard robotic training, movements are visually guided through gamified computer interfaces(*2*). From studies with unimpaired individuals, we know that the motor system combines feedback from vision and proprioception to create an internal model of the limb that it uses for control(*33, 34*). When proprioception is impaired, the motor system likely increases reliance on visual feedback to compensate for (but not retrain) proprioceptive deficits. To more directly engage and challenge proprioception during training, the Propriopixel mode replaces visual gaming elements with proprioceptive cues provided by the robot, requiring the creation of motor plans based on perceived proprioceptive input(*35*). We recently showed that this approach was feasible for individuals with impaired finger proprioception, who found it highly motivating(*36*).

We also developed a second proprioception-focused paradigm, Virtual Assistance mode, which does not provide physical assistance but remains engaging(*36*). Prior studies have shown that visually occluded, unassisted movement training can improve proprioception(*23, 37*), while similar training performed with physical assistance may not(*19*). It has been proposed that physical assistance introduces a "credit assignment" problem, where participants have difficulty distinguishing the sensory consequences of their own motor actions from those generated by the assistive forces, which hinders sensorimotor learning(*38, 39*). However, physical assistance is a key means of enhancing gameplay success, which increases motivation and stimulates reward circuitry that contributes to motor learning(*28, 30, 40*). To ameliorate the credit assignment issue and maintain comparable motivational support, we developed a Virtual Assistance paradigm that dynamically adjusts game parameters to maintain high-levels of training success comparable to standard robotic training without providing physical assistance(*36*). This approach enables a controlled comparison of whether removing physical assistance can improve proprioceptive learning and sensorimotor recovery, by increasing the focus on self-generated proprioceptive cues without externally generated proprioceptive feedback.

To test the relative efficacy of these new training modes on hand movement recovery, participants in the chronic phase of stroke (N=46) participated in 3 weeks of movement training (nine 2-hour sessions) using the FINGER robot(*41, 42*). They were randomized into one of three training groups: Standard,

Propriopixel, or Virtual Assistance. The primary outcome measure was change in UE function, measured with the Box and Block Test (BBT) from baseline to one month follow-up. Secondary outcomes included a novel electroencephalogram (EEG) measure that captures neural estimation of passive finger movement, which we used to understand how these training modes alter proprioceptive processing. We hypothesized that individuals with impaired proprioception would experience greater gains in hand function and proprioceptive ability from the proprioceptively-focused training modes.

## RESULTS

Chronic stroke participants (N=46) completed three weeks of finger and thumb movement training with the FINGER robot in one of three randomly allocated paradigms: Standard (visual gaming with physical assistance), Propriopixel (propriopixel gaming with physical assistance), or Virtual Assistance (visual gaming with virtual assistance). Both Propriopixel and Virtual Assistance modes were designed for individuals with impaired proprioception, to either enhance attention to proprioception during training (Propriopixel), or to remove potentially confusing, externally driven proprioceptive feedback (Virtual Assistance). The primary outcome was change in hand function (Box and Block Test, BBT) from baseline to one-month follow-up (1MFU), with secondary outcome measures including additional clinical, proprioceptive, and EEG-based measures (Tbl. S1).

Participant demographics and baseline clinical scores are reported in Table 1. There were no significant differences between training groups' BBT scores at baseline nor for any other metric. The BBT score did not change significantly between the two baseline evaluations (paired t-test, p=0.86), which were an average of 6 ± 2 days apart, confirming a stable baseline.

We defined individuals as having significant finger proprioceptive impairment if their Crisscross error exceeded two standard deviations above the mean of an unimpaired, age-matched control group (N=37, aged: 55.6 ± 17.6 years, mean error ± std: 7.68 ± 2.56 deg). Twenty (44%) chronic stroke participants met this criterion (Standard Training: N=5; Propriopixel Training: N=8; Virtual Training: N=7).

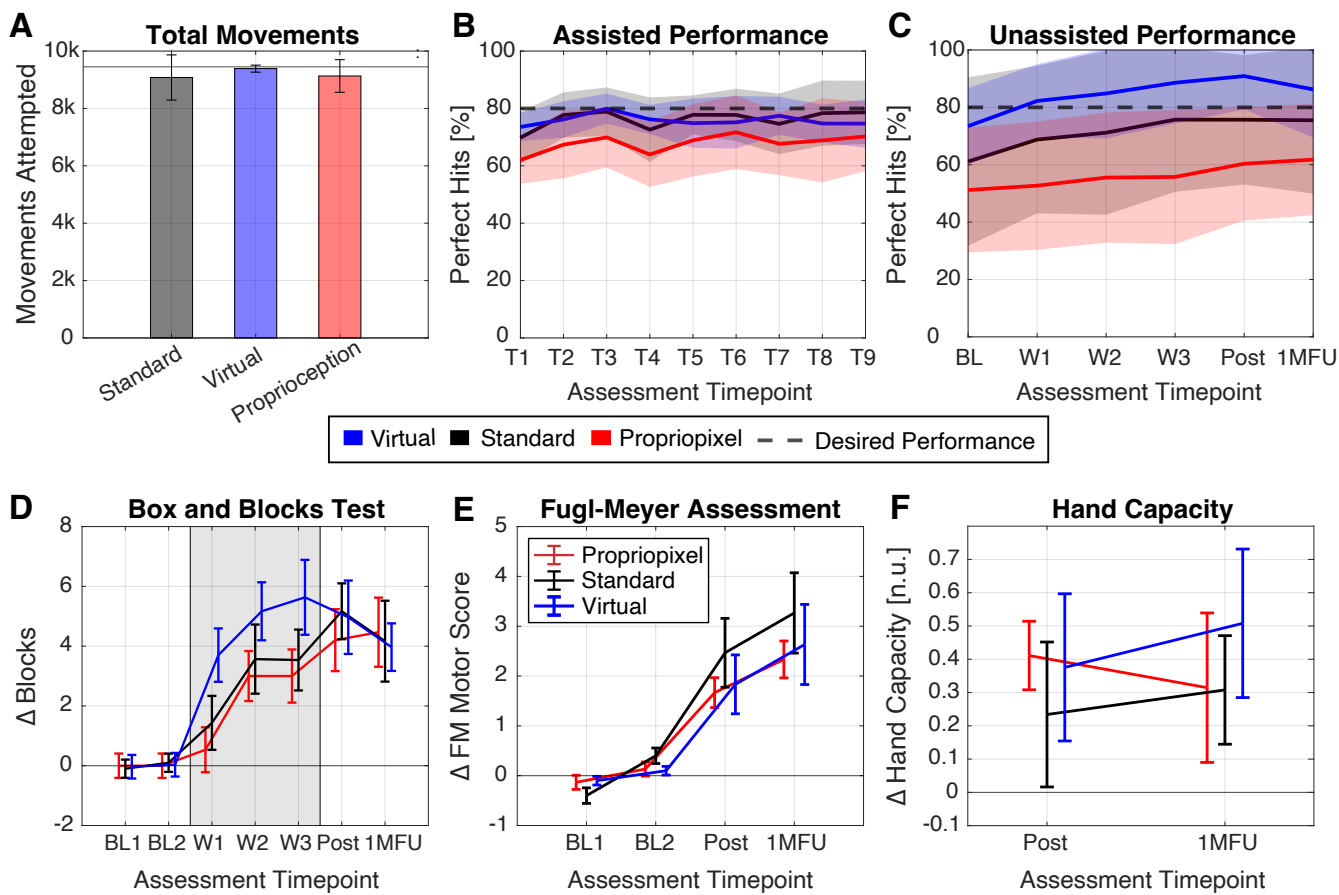

*Figure 1: Training metrics and outcome measures. Figures report the mean and error bars/shaded regions represent the standard error of the mean. T1-T9 are therapy sessions; W1-W3 are weeks. BL=baseline, Post=post therapy assessment, 1MFU=1 month follow-up assessments. A: Total number of movements attempted across both training games in each group. B. Assisted gameplay success (averaged across games); assistance gains were adjusted in sessions 1, 4, and 7. Propriopixel success was slightly lower than visually guided modes, significant only in week 1. C: Unassisted gameplay success averaged across games. The Virtual and Standard groups trained in the same unassisted, visually guided interface, while the Propriopixel group used an unassisted Propriopixel mode, which was more challenging and yielded lower success rates. D: Change in number of blocks transferred during BBT from the average baseline value. E: Change in Fugl-Meyer Assessment of Upper Extremity Motor status from the average baseline value. F: Change in Hand Capacity, a measure of finger strength and individuation, from the baseline value.*

|  | All (N=45) | Standard (N=15) | Virtual (N=15) | Propriopixel (N=15) | Effect of Group |  |
|---|---|---|---|---|---|---|
| **Demographics** | median [IQR] | median [IQR] | median [IQR] | median [IQR] | p, F(DF) | - |
| Age (years) | 60 [46-68] | 58 [41-69] | 52 [46-70] | 62 [59-68] | 0.44, 1.6(2) | - |
| Time Since Stroke (months) | 46 [20-80] | 67 [33-94] | 40 [10-66] | 49 [24-64] | 0.21, 3.1(2) | - |
| Gender [Male(M)/Female(F)] | 33M/12F | 11M/4F | 12M/3F | 10M/5F | 0.72, 0.67(2) | - |
| Handedness [Right(R)/Left(L)] | 40R/3L | 13R/2L | 13R/1L | 14R/0L | 0.16, 3.6(2) | - |
| Side of Hemiparesis [Right(R)/Left(L)] | 20R/25L | 6R/9L | 7R/8L | 7R/8L | 0.92, 0.18(2) | - |
| Type of Stroke [Ischemic (I)/ Intracerebral Hemorrhagic (H)] | 25I/19H | 8I/7H | 6I/8H | 11I/4H | 0.16, 3.7(2) | - |
| **Baseline Measures** | median [IQR] | median [IQR] | median [IQR] | median [IQR] | p, F(DF) | Baseline 2-Baseline 1 Effect size, Median [IQR] |
| Box and Blocks Test [blocks, max 150] | 24 [11-38] | 16 [5.6-41] | 25 [16-34] | 22 [7.5-38] | 0.81, 0.42(2) | -0.04, 1 [-2,2] |
| Fugl-Meyer Upper Extremity Motor Score [ max 66] | 50 [36-56] | 44 [30-55] | 56 [45-57] | 50 [37-54] | 0.053, 5.9(2) | **0.43, 0 [0,1]** |
| Fugl-Meyer Upper Extremity Sensory Score [max 20] | 20 [18-20] | 20 [14-20] | 20 [18-20] | 20 [19-20] | 0.81, 0.41(2) | -0.01, 0 [0,0] |
| Motor Activity Log- Quality [max 5] | 1.5 [0.95-2.7] | 2.1 [1.2-2.8] | 1.9 [0.86-3] | 1.4 [0.65-2.2] | 0.32, 2.3(2) | - |
| Motor Activity Log- Quantity [max 5] | 1.5 [0.88-2.7] | 2.2 [0.99-2.8] | 2.1 [0.97-3] | 1.2 [0.53-2.1] | 0.21, 3.1(2) | - |
| 9 Hole Peg Test [time, s] | 60 [43-60] | 60 [44-60] | 60 [41-60] | 60 [45-60] | 0.87, 0.27(2) | **-0.39, 0 [-4,0]** |
| Trail Making Test A and B [time, s] | 124 [98.4-173] | 133 [90.4-174] | 117 [101-187] | 124 [98.8-164] | 0.98, 0.049(2) | - |
| Modified Ashworth Spacticity Scale [max 10] | 0 [0-0.45] | 0.1 [0-0.95] | 0 [0-0] | 0 [0-0.7] | 0.059, 5.7(2) | - |
| Visual Analoge Pain Scale | 0 [0-0] | 0 [0-0] | 0 [0-0] | 0 [0-0] | 0.56, 1.2(2) | -0.3, 0 [0,0] |
| Hand Capacity [n.u.] | 1 [0.12-1.9] | 0.52 [0.094-1.3] | 1.7 [0.82-2.1] | 0.31 [0.077-1.8] | 0.1, 4.6(2) | - |
| Crisscross [error, deg] | 12 [7.2-17] | 7.4 [6.1-19] | 11 [7.8-18] | 14 [9-17] | 0.34, 2.1(2) | -0.02, 0 [-1,1] |
| Move and Match [error, deg] | 7.9 [5.2-9.8] | 6.5 [4.9-9.2] | 6.1 [5.1-10] | 8.9 [5.2-10] | 0.72, 0.64(2) | -0.02, 0 [-1,1] |
| ThumbSense [error, % missed] | 15 [7.5-33] | 12 [7.5-32] | 18 [5-28] | 20 [10-40] | 0.51, 1.4(2) | -0.004, 0 [-10,10] |

*Table 1: Baseline demographics and clinical/robotic assessment scores. No significant differences were observed between groups, although Fugl-Meyer Upper Extremity scores approached significance between Virtual and Standard groups (p=0.053), and MASS spasticity was slightly higher in the Standard training group (p=0.059), but was low overall. For measures with two baseline assessments (right-most column), FM Motor and 9HPT scores showed small yet significant (p<0.05) improvements between baseline 1 and 2 (bolded).*

**Training metrics**

Participants attempted an average of 9,200 ± 568 movements over three weeks, with no significant group differences (Standard: 9,079 ± 786; Virtual: 9,389 ± 122; Propriopixel: 9,132 ± 568; Wilcoxon $\chi^2(2)$=4.38, p > 0.11; Fig. 1A). Average assisted game success was 82.4% ± 6.2% (Standard), 80.1% ± 4.7% (Virtual), and 74.0% ± 11.7% (Propriopixel). Assisted success rates did not differ from the target 80% (t-test, p > 0.07), though the Propriopixel group achieved significantly lower success overall (Group effect: F(2)=4.71, p=0.012; Tukey: Propriopixel vs. Virtual, p<0.015; vs. Standard, p<0.047). Post hoc analysis showed lower success was due to greater timing errors (early/late hits) but not missed movements. Timing errors improved across sessions with training, such that after week 1, no group differences remained (Fig. 1B). Thus, we considered groups well matched across movements performed and success rates achieved.

| Change at 1 Month Follow Up | All (N=45) | Standard (N=15) | Virtual (N=15) | Propriopixel (N=15) | Group | Time Point | 1MFu X Group |
|---|---|---|---|---|---|---|---|
| Clinical Outcomes | median [IQR] | median [IQR] | median [IQR] | median [IQR] | p, effect(DF) | p, effect(DF) | p, effect(DF) |
| Box and Blocks Test [blocks, max 150] | **3 [1.5, 6.1]** | **2.5 [0.5, 7.4]** | **3 [2.6, 5.2]** | **3 [2, 7.1]** | 0.84, 0.34(2) | **<0.00001, -5.3(1)** | 0.55, 1.2(2) |
| Fugl-Meyer Upper Extremity Motor Score [max 66] | **2.5 [1, 4]** | **4 [1.2, 5.1]** | **2 [0.25, 2.9]** | **2.5 [1.1, 3.5]** | 0.063, 5.5(2) | **<0.00001, -5.2(1)** | 0.5, 1.4(2) |
| Fugl-Meyer Upper Extremity Sensory Score [max 20] | 0 [0, 0] | 0 [0, 0] | 0 [0, 0] | 0 [0, 0] | 0.7, 0.71(2) | 0.079, -1.4(1) | 0.6, 1(2) |
| Motor Activity Log- Quality [max 5] | **0.58 [-0.055, 1.1]** | **0.74 [0.084, 1.1]** | **0.87 [0.28, 1.2]** | 0.19 [-0.27, 0.82] | 0.17, 3.5(2) | **<0.00001, -3.9(1)** | 0.19, 3.3(2) |
| Motor Activity Log- Quantity [max 5] | **0.39 [0.12, 1.1]** | **0.68 [0.12, 0.86]** | **0.76 [0.33, 1.3]** | 0.16 [-0.22, 0.83] | 0.12, 4.3(2) | **<0.00001, -4.6(1)** | 0.21, 3.1(2) |
| 9 Hole Peg Test [time, s] | **-15 [-32, 3.9]** | **-20 [-31, 5]** | **-20 [-36, 7.8]** | **-13 [-24, -7]** | 0.94, 0.11(2) | **0.0032, 2.7(1)** | 0.96, 0.08(2) |
| Trail Making Test A and B [time, s] | 0 [-2.2, 0] | 0 [-3, 0] | 0 [-2.2, 0] | 0 [-1.4, 0] | 0.89, 0.24(2) | **0.002, 2.9(1)** | 0.95, 0.11(2) |
| Modified Ashworth Spacticity Scale [max 10] | 0 [0, 0] | 0 [0, 0.075] | 0 [0, 0] | 0 [0, 0.2] | **0.019, 7.9(2)** | 0.18, -0.93(1) | 0.46, 1.6(2) |
| Visual Analoge Pain Scale | 0 [0, 0] | 0 [0, 0] | 0 [0, 0] | 0 [0, 0] | 0.53, 1.3(2) | 0.43, 0.17(1) | 1, 0.0034(2) |
| Robotic Outcomes | median [IQR] | median [IQR] | median [IQR] | median [IQR] | p, effect(DF) | p, effect(DF) | p, effect(DF) |
| Hand Capacity [n.u.] | **0.06 [-0.05, 0.75]** | **0.04 [-0.01, 0.29]** | 0.2 [-0.06, 0.85] | 0.06 [-0.09, 0.88] | 0.26, 2.7(2) | **0.004, -2.6(1)** | 0.91, 0.18(2) |
| Crisscross [error, deg] | **-0.89 [-3.5, 0.5]** | -0.63 [-1.9, 0.5] | -0.84 [-3.9, 0.52] | **-3.5 [-4.1, 0.53]** | 0.19, 3.4(2) | **0.0027, 2.8(1)** | 0.29, 2.4(2) |
| Move and Match [error, deg] | **-0.8 [-1.3, 0.14]** | 0.14 [-0.47, 1.7] | -0.79 [-1, -0.066] | **-1.9 [-2.9, -0.85]** | 1, 0.0057(2) | **0.011, 2.3(1)** | **0.0029, 12(2)** |
| ThumbSense [error, % missed] | -5 [-12, 0] | -2.5 [-14, 11] | **-7.5 [-15, -2.5]** | -3.8 [-10, 0] | 0.63, 0.94(2) | **0.013, -2.2(1)** | 0.68, 0.76(2) |

*Table 2: Analysis of change in clinical and robotic outcome measures of sensorimotor hand function. Main effects of timepoint (baseline vs. 1-month follow-up) were assessed with Wilcoxon sign-rank testing. Group differences across timepoints and training × time interactions were evaluated using Kruskal-Wallis tests on change scores. We used post-hoc Wilcoxon signed rank testing to determine within group effects (one-tailed) and Wilcoxon rank sum testing to determine between group effects (two-tailed). Significant results (p<0.05) are bolded. Expanded results are reported in the supplemental materials (Table S3, Figure S2).*

**Improvements in hand movement ability with training**

Participants showed significant improvements in BBT scores and several secondary clinical outcomes (Fig. 1D, Tbl. 2), including upper extremity motor impairment (Fugl-Meyer Assessment, UE Motor, Fig. 1E) and self-reported use of the impaired hand at home (Motor Activity Log). They also improved on the Trail Making A and B tests, which assess processing speed, visual attention, and cognitive flexibility (Tbl. 2). Consistent with baseline results, the Standard training group had slightly higher spasticity, but it was negligibly low across groups (Tbl. 2, p<0.019). No other significant effects of group or group × timepoint interactions were observed for any clinical outcome.

Participants' unassisted gameplay success improved over time (Friedman's test, $\chi^2=5.16(1)$, $p<0.0001$), starting from the second week of training onward (Wilcoxon ranksum, $p<0.0231$, Fig. 1C). There was a significant effect of group (Kruskal Wallis, $\chi^2(2)=17.01$, $p<0.0002$), driven by the Propriopixel group having lower success across all timepoints. However, there was no significant difference between groups change with training (Kruskal Wallis, $\chi^2(2)=2.69$, $p>0.26$), indicating that all groups had similar improvement from baseline assessments. Hand capacity(*43*), which reflects finger strength and individuation, also improved (Wilcoxon signed rank, $p<0.004$, Tbl. 2, Fig. 1F). No significant effects of group or group × timepoint interactions were observed for hand capacity.

These results suggest comparable improvements across training conditions (at the group level) in hand movement ability, quantified robotically or clinically.

**Improvements in hand movement ability related to baseline finger proprioception ability**

Baseline finger proprioception, measured by the Crisscross assessment, predicted responsiveness to Standard training, quantified by change in BBT (Fig. 2A). Participants with more intact proprioception showed greater immediate post-therapy improvement ($R^2(15)=0.57$, $p=0.002$), with a similar trend at 1MFU ($R^2(15)=0.12$, $p=0.21$). The Propriopixel training group showed an opposite trend at 1MFU ($R^2(14)=0.13$, $p=0.20$), while the Virtual Assistance group showed no association to baseline proprioception ($R^2(14)=0.00$, $p=0.88$, Fig. S3).

To complement this correlational analysis, we subdivided each training group into individuals with or without baseline finger proprioceptive impairment based on their baseline Crisscross error (detailed above). For the Standard group, only four participants met the impairment criterion. To increase power, we included data from 15 participants from a prior FINGER trial who satisfied the same inclusion criteria, performed the same Crisscross assessment, and received comparable training (Methods). This yielded a combined Standard training group with 12 impaired and 18 unimpaired participants.

Within the combined Standard training group, participants with impaired proprioception had smaller gains in BBT at 1MFU than those with intact proprioception (Fig. 2B, Kruskal-wallis, $\chi^2(1)=6.2$, $p<0.013$).

The Propriopixel training group showed the opposite pattern; participants with impaired proprioception tended to have larger BBT gains at 1MFU than those with intact proprioception (Fig. 2B, $\chi^2(1)=4.29$, $p<0.058$). Between training groups, individuals with impaired proprioception who received Virtual or Propriopixel training experienced larger increases in BBT compared to those who received Standard training (Kruskal-wallis, $\chi^2(2)=11.97$, $p<0.0025$). For these individuals, Propriopixel training improved BBT scores significantly more than Standard training (Wilcoxon ranksum: $p<0.0028$), with Virtual training trending similarly ($p<0.068$). These differences in BBT response could not be attributed to differences in baseline BBT score between sub-groups (see Supplemental Materials for details).

Together, these findings indicate that participants with proprioceptive impairment benefited more from Propriopixel and Virtual Assistance training, whereas those with intact proprioception responded best to Standard training.

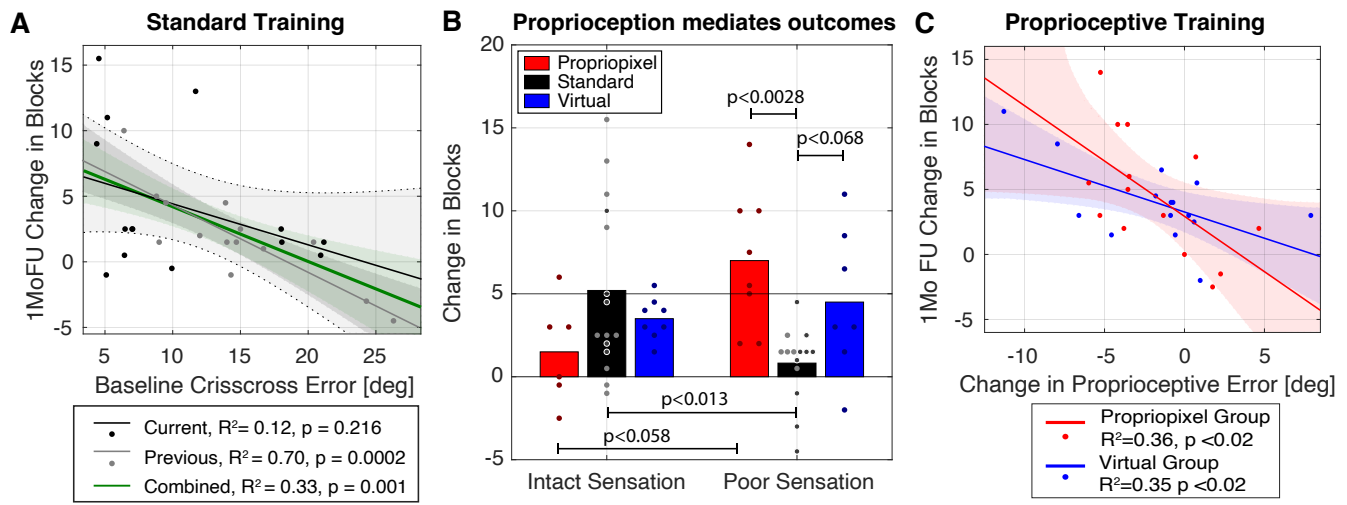

*Figure 2: Proprioceptive outcomes and their relationship to hand function change A: Greater baseline proprioceptive ability predicts greater improvement in BBT following Standard robotic training. Data includes participants from the current study and a matched training group from a previous randomized clinical trial using FINGER. B: Training response depended on baseline Crisscross proprioception ability. Individuals with impairments (> 2SD normal) improved more in Propriopixel and Virtual training modes compared to Standard training. C: Improvements in proprioception, quantified by Crisscross, were significantly related to improvements in BBT in the Propriopixel and Virtual training groups.*

**Improvements in hand sensory ability with training**

A key question was whether hand sensory ability improved with training, particularly with the proprioceptive-focused training techniques. The FM-Sensory score, a coarse clinical scale, showed no significant change at the group level (Tbl. S4). Results of the robotic assessments, including Crisscross, Move and Match and ThumbSense (Tbl. 2), provide greater resolution of hand sensory ability(*2*).

Crisscross error decreased significantly for all participants across training (median [IQR]: -0.89 [-3.5, 0.55] deg, Wilcoxon signed-rank: $p<0.0055$), with a trend towards greater improvements in the Propriopixel and Virtual training groups compared to Standard training (Kruskal-Wallis, $\chi^2(1)=1.99$, $p=0.16$). Within group analysis showed significant improvements in proprioceptive ability following Propriopixel (-3.5 [-4.09, 0.52] deg, $p<0.029$) and Virtual training (-0.84 [-3.9, 0.52] deg, $p<0.015$), but not Standard Training (-0.63 [-1.9, 0.5] deg, $p=0.195$).

Participants with proprioceptive impairment (Wilcoxon signed-rank, $Z(20)=2.41$, $p<0.026$) and those in the normal, age-matched range ($Z(25)=1.64$, $p<0.05$) improved their Crisscross performance by a similar degree (Wilcoxon rank-sum, $p>0.70$, all participants: 9.5% [-5.0, 28.9]%, impaired: 8.51% [-2.6, 30.0]%, unimpaired: 9.4% [-7.6, 28.8]%).

Move and Match tracking error also improved across groups at 1MFU (Wilcoxon signed-rank: -0.8 [-1.8, 0.14] deg, $p<0.011$). However, change in performance significantly differed between groups (Kruskal-Wallis, $\chi^2(2)=11.71$, $p<0.003$; Propriopixel vs Standard: $p<0.0035$). Only the Propriopixel training group significantly reduced tracking error following training (-1.9 [-2.9, -0.85] deg; $p<0.001$), while the Virtual group neared significance (-0.79 [-1, -0.07] deg, $p<0.07$), and Standard training had no change (0.14 [-0.47, 1.7] deg, $p=0.29$). Move and Match tracking error was weakly correlated with Crisscross error before training ($R^2_{adj}(34)=0.13$, $p<0.031$), an association that increased following training ($R^2_{adj}(32)=0.34$, $p<0.0003$).

Thumb proprioceptive ability measured with ThumbSense also improved at the group level (Wilcoxon signed rank: 5 [0,12]%, $\chi^2(1)=2.2$, $p<0.013$), with no significant difference between groups (Kruskal-Wallis, $p > 0.68$). Within group, both the Virtual (7.5 [2.5, 15]%, $p<0.042$) and Propriopixel (3.8 [0, 10]%,

p<0.031) groups improved in ThumbSense performance, while the Standard training group did not (-2.5 [-11, 14]%, p=0.33).

The observed decreases in proprioceptive error were associated with improvements in BBT score for the Propriopixel group (Fig. 2C, $R^2(14)=0.35$, p<0.021) and in the Virtual Assistance group (Fig. 2C, $R^2=0.35$, p=0.02). There was no relationship between change in proprioception error and change in BBT in the Standard training group ($R^2(14)=0.02$, p<0.65).

Together, these findings indicate that proprioceptively-focused training, especially Propriopixel training, improved proprioceptive ability, and improvement were associated with hand function recovery.

**Improvements in a novel neural marker of proprioceptive processing**

We sought to gain insight into the neural correlates of finger proprioception during the Crisscross assessment. During EEG assessment, participants performed 6 runs of the Crisscross assessment with performance feedback to increase sustained task engagement (see methods for more details). Our analysis focused on the Contingent Negative Variation (CNV), a slow negative event-related potential that reflects anticipatory processing between an initial warning stimulus and a subsequent imperative stimulus that cues an action(*44*, *45*). In the Crisscross task, sensed movement onset is the warning stimulus, and the perceived finger crossing is the imperative stimulus to push the response button. To our knowledge this is the first study of the CNV observed in response to purely proprioceptive cues (Fig. 3A). We therefore refer to it as a proprioceptive CNV (pCNV) and examined its relationship to baseline proprioceptive ability and training related changes.

At baseline, larger pCNV deflections (electrodes Fz,F3,Cz,C3,Pz,P3) were associated with better Crisscross performance (Fig. 3B, $R^2(41)=0.322$, p<0.0001), an association that increased in strength at follow-up ($R^2(43)=0.541$, p<0.00001). Further, participants who improved their Crisscross performance (3-50% error reduction) had increased pCNV deflections following training (Wilcoxon-signrank test, Z(29)=2.24, p<0.025), while individuals who got worse at the assessment (2.4-60% error increase) did not (Wilcoxon-signrank test, Z(11)=-0.89, p<0.38, Fig. 3C). Thus, pCNV performed as a biomarker of proprioceptive ability, both in relation to baseline status and change with training.

We further tested for differences in pCNV response between training groups (Fig. 3C). At baseline, there were no differences between training groups in task performance (Kruskall-Wallis, p>0.25) or in pCNV magnitude (Kruskall-Wallis, p>0.43). Following training, the Virtual and Propriopixel training groups significantly reduced their Crisscross error (p<0.045), while the Standard training group did not (p>0.18), in line with Crisscross assessment (without feedback) results (Fig. S4). Following training, there was a significant difference between training groups pCNV response measured in the Pz electrode (Kruskal Wallis, p=0.038); the proprioceptive-focused training groups had greater increases in pCNV deflections with training compared to the Standard Training group (Fig. 3C.2, Wilcoxon ranksum test, Combined proprioceptive: p<0.018; Propriopixel: p<0.017; Virtual: p=0.085). Localization of effects to Pz (located over the parietal cortex) suggest a somatosensory cortex origin.

To identify which Crisscross task feature best explained pCNV modulation, we performed a correlational analysis between task parameters (finger position, velocity, error, see Fig. S6) and the measured EEG time series across the whole task. Finger velocity yielded the strongest associations in the same topographical regions as the pCNV (Fig. 3E) and similarly increased in strength with improved Crisscross performance. Following training, the correlation between velocity and activity measured in sensorimotor electrodes (Fig. 3F, P3, Pz, Cz) increased in the proprioceptive- focused training groups to a greater extent than the Standard training group (Kruskal Wallis, p=0.021; Combined proprioceptive: -0.044 [-0.89, -0.014]; Standard: 0.01 [-0.028, 0.033]). Again, the Pz electrode showed the greatest difference between training groups (Kruskal Wallis, p=0.037; Propriopixel: -0.066 [-0.11,-0.007]; Virtual: -0.039 [-0.08,0.023]; Standard: 0.023 [-0.026,0.045], Propriopixel vs Standard: p<0.021).

These results suggest that the proprioceptive-focused training modes, especially Propriopixel training, enhanced neural sensitivity to proprioceptive cues, particularly estimation of finger velocity.

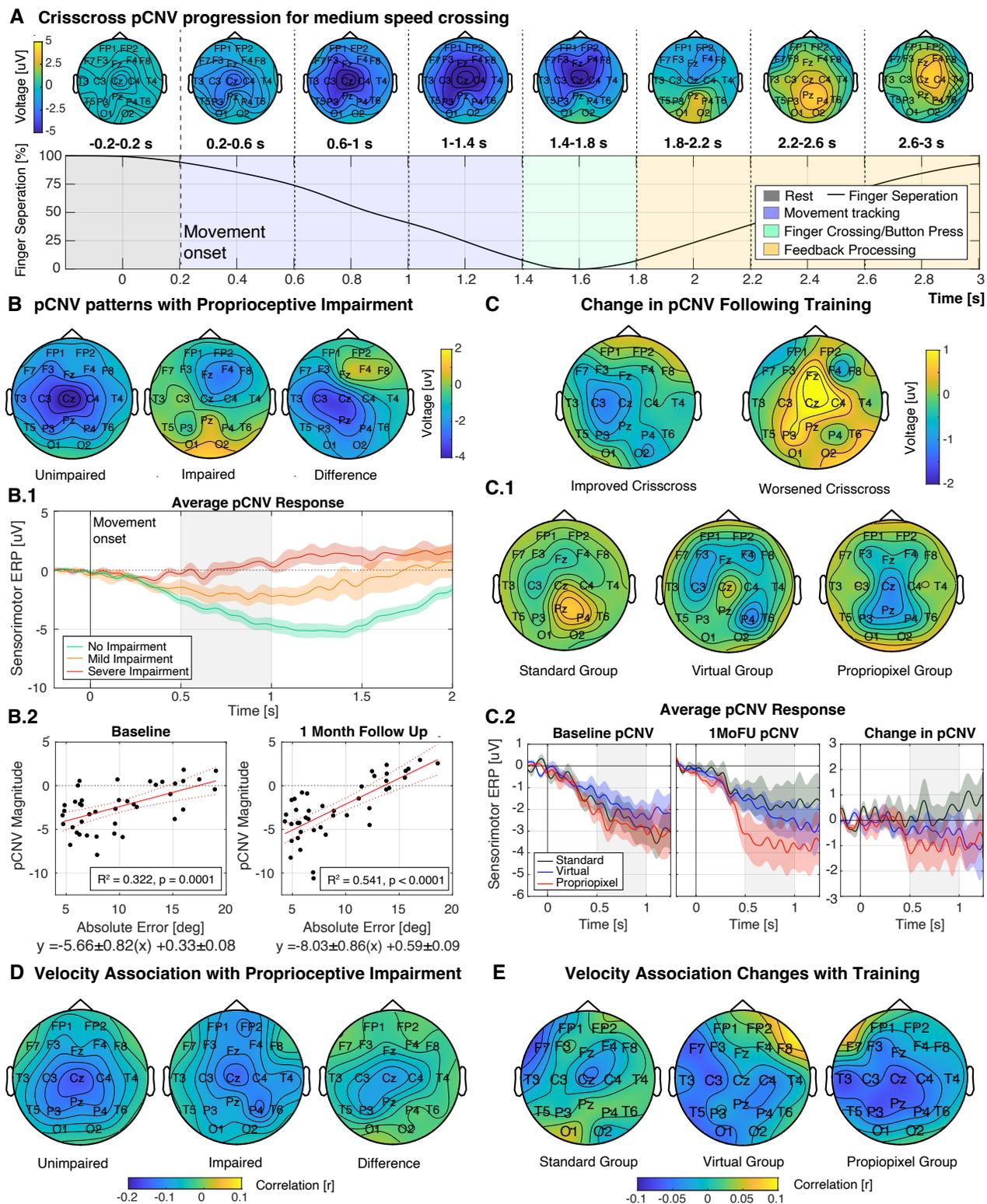

*Figure 3: The proprioceptive Contingent Negative Variation (pCNV) measured during the Crisscross proprioception task. Data were adjusted for affected side so left-hemisphere electrodes correspond to the affected hand in the robot. Participants were classified as impaired/unimpaired by comparison to an age-matched control group (N=36). Sensorimotor electrodes include P3/Pz, C3/Cz and F3/Fz. **A:** Average pCNV for individuals with unimpaired*

*proprioception during medium-speed trials. The negative deflection begins just after finger movement onset (time=0), peaks near the button press (and finger crossing), and rebounds positively during feedback processing. Fig. S5 reports response for slow and fast speed trials. **B:** Topographical maps show smaller pCNV over left posterior sensorimotor electrodes in impaired individuals, indicating reduced sensory cortex engagement. **B.1:** pCNV magnitude over sensorimotor electrodes varied with impairment, binned into tertiales based on crisscross performance. **B.2:** Correlation between pCNV magnitude (0.5–1 s post-movement) and Crisscross performance strengthened following training. **C:** Training increased pCNV magnitude over sensorimotor regions, with greater negative deflections associated with improved performance. **C.1–C.2:** Both Virtual and Propriopixel groups showed significant increases, largest in Propriopixel. **D:** Correlation between finger velocity and neural activity mirrored pCNV topography and varied with proprioceptive impairment. **E:** Propriopixel training yielded the greatest increases in velocity-related neural association (Cz,Pz,P3), followed by Virtual training.*

## DISCUSSION

We evaluated two novel robotic training paradigms designed for individuals with proprioceptive deficits and compared them to standard, visually-driven, physical assist-as-needed robotic training. At baseline, 44% of the participants had impaired finger proprioception. These individuals showed significantly greater improvements in hand function, measured by the Box and Blocks Test (BBT), and in proprioceptive ability when trained with the proprioceptively-focused forms of training compared to standard robotic training. Improvements in hand function were associated with improvements in finger proprioception. Further, we identified a novel EEG-based neural marker, the proprioceptive CNV (pCNV), that reflects the process of estimating finger movement speed during proprioceptive testing. Individuals with poorer proprioception had reduced pCNV magnitude, which improved with training, particularly for those who received proprioceptively-focused training. These findings have specific implications for treatment of sensorimotor deficits after stroke and have broader implications for precision rehabilitation.

**The role of proprioception in shaping the response to standard robotic training**

Proprioception is a primary sensory channel for online feedback control of limb movement, and plays a key role in the trial-by-trial learning of new movements(*46, 47*). In Standard robotic training, physical assistance augments proprioceptive input by increasing training range of motion and movement accuracy(*11–13*). For individuals with intact proprioception, identified by robotic assessment of passive finger movement sense, this type of training is effective: participants in this subgroup significantly improved their hand function, consistent with previous outcomes following standard robotic training(*19*). However, for individuals with impaired finger proprioception standard robotic training was consistently ineffective. Strikingly, in both the prior(*19*) and current clinical trials, not a single participant with proprioceptive impairment, achieved a clinically meaningful improvement in BBT score (an increase of ≥6 blocks(*48, 49*).

These findings suggest that intact proprioception is necessary to interpret and integrate the enhanced somatosensory feedback provided by physical assistance, promoting effective engagement of sensory feedback–dependent mechanisms underlying motor learning. This replication of findings strengthens the evidence that proprioceptive integrity, specifically related to passive movement sense, is a key moderator of training effectiveness and underscores the need for tailored movement training strategies according to degree of proprioceptive impairment.

**Proprioceptively-focused strategies benefited individuals with impaired proprioception**

Previous proprioceptively-focused robotic training approaches used vibrotactile feedback(*21, 50, 51*), and/or error amplification of small movements(*23, 52*) to enhance sensory feedback and attention to movement errors during training. While these methods provide some benefit, they augment feedback rather than directly retrain proprioceptive ability.

Here, we designed a novel training technique, Propriopixel training, that required participants to actively and continuously attend to proprioceptive cues of self-generated and robot facilitated movements to form motor plans for playing the rehabilitation games. Participants with impaired proprioception who received Propriopixel training achieved substantial gains in BBT score that were significantly higher than those who received Standard training. These participants demonstrated marked improvement in their finger proprioceptive deficits that were significantly associated with gains in hand motor function. They also exhibited significant gains in pCNV magnitude, associated with improvements in neural estimation of passive finger movement in the Crisscross assessment.

From these results, we hypothesize that the emphasis on proprioceptive perception during training reconditioned attentional mechanisms to increase sensorimotor signaling and integration, which in turn promoted improved higher-order processing of proprioceptive input for motor planning. These improvements in proprioception may have led to better feedback control of the hand, improving BBT and other scores of motor ability. Alternatively, enhanced sensorimotor processing may have improved the quality of the proprioceptive teaching signal needed to reorganize cortical motor circuits during training, facilitating greater motor learning that led to improved hand function. These concepts align with evidence that sensory attention tasks transiently modulate the integration of somatosensory input into both sensory and motor cortex, and this modulation can enhance subsequent motor learning(*53–55*).

We also tested a Virtual Assistance paradigm, which aimed to enhance proprioceptive processing by removing physical assistance, thus eliminating potentially confounding effects of robotic assistance on interpretation of proprioceptive input. In this paradigm, participants were required to proprioceptively monitor self-generated movements to accurately transform visual information provided by the games into appropriate motor commands. Participants in this group significantly improved their hand function, finger proprioception, and pCNV, and improvements in proprioception were associated with gains in hand function similarly to Propriopixel training. For individuals with proprioceptive deficits, Virtual Assistance training also yielded better outcomes than Standard Training.

These findings suggest that eliminating physical assistance removed distracting external sources of proprioceptive input and/or heightened attention to self-generated proprioceptive information. Although Virtual Assistance training focused on proprioceptive estimation of self-generated movement, training significantly improved passive movement estimation in the Crisscross assessment suggesting some overlap in estimation of self-generated and externally applied movements. This overlap is also reflected in the increased association between performance in the Crisscross and the Move and Match assessments following training, indicating robotic training modulated some shared neural mechanism. These findings are consistent with prior studies that have shown unassisted movement practice can improve proprioception(*23*), and that proprioceptive learning exhibits some transfer between active and passive modalities(*50, 56*).

Our findings align with prior work that demonstrated proprioception can be trained in healthy individuals and retrained in a variety of neurologic conditions, including stroke(*23, 50, 56, 57*). Consistent with these results, we found that training techniques that emphasize proprioceptive awareness during motor planning are generally more effective than generalized movement strategies(*50, 56*). Our results extend previous findings by linking these sensory improvements to clinically meaningful gains in motor performance in chronic stroke, and to improvements in conscious processing of proprioceptive information.

**Insights into proprioceptive processing and its plasticity with robotic training**

To investigate the neural mechanisms underlying passive movement sense, we examined the contingent negative variation (CNV) as a potential marker of central proprioceptive processing during Crisscross, a proprioceptively guided task. The CNV is a slow negative cortical potential that reflects anticipatory attention and motor preparation between an initial and a subsequent cue. In tasks using audio-visual cues, reduced CNV amplitude and delayed latency have been linked to impaired motor planning and attention after stroke(*58–60*). Recently, the CNV response was linked to impaired attention-mediated sensorimotor processing in individuals with functional movement disorders (FMD) performing a visually-cued motor task(*45*). The CNV response was initially absent in individuals with FMD but normalized following physical therapy, paralleling clinical motor recovery, suggesting the CNV may capture the functional coupling between attentional engagement and improved sensorimotor control.

Here, we extended this work to examine the CNV in a novel proprioceptive context. In Crisscross, the initial cue is the sensed onset of passive finger movement and the second cue is the sensed crossing of the fingers. Thus, the resulting response must be proprioceptively driven (a proprioceptive CNV, pCNV). To our knowledge, this is the first demonstration of a pCNV.

The pCNV magnitude was reduced in individuals with impaired proprioceptive processing but increased after training. The largest gains occurred in the Propriopixel group, which also showed the greatest improvements in proprioceptive performance. Unlike the traditional CNV, which localizes to midline electrodes (Fz/Cz), the pCNV lateralized over sensorimotor regions of the hand being passively moved (C3, P3) and the posterior parietal cortex (Pz). The pCNV magnitude was decreased in individuals with impaired proprioceptive processing and improved with training. Its magnitude covaried with passive finger movement velocity, consistent with prior findings implicating these regions in online proprioceptive processing of passive finger and wrist motion (Fig. 3A,E)(*61–64*). The primary somatosensory cortex (underlying P3) processes afferent position and movement signals(*65*), and is interconnected with the posterior parietal cortices (underlying Pz), which integrate visual, tactile, and proprioceptive information to inform voluntary action(*66, 67*). In our cohort, the strongest training- and impairment-related differences in pCNV magnitude were observed at the electrodes overlying these regions (Pz, P3).

We interpret these findings to support the pCNV as a measure of parietal-based anticipatory attention that mediates higher order processing of afferent proprioceptive information. Attentional networks act as a filter that gate the extent to which proprioceptive information is consciously processed. An extreme example of this is neglect(*53*), in which attentional deficits produce a functional sensory loss even though afferent sensory tracts are intact. In our cohort, it appears that proprioceptively-focused training improved participants ability to consciously attend to and process proprioceptive input, resulting in an increased pCNV signal magnitude and improved crisscross performance.

Together, these findings highlight the pCNV as a promising neurophysiological biomarker for attentional processing of proprioception, with implications for both mechanistic understanding and a tailored therapeutic approach to rehabilitation therapy. The data suggest that training paradigms that increase attentional demands on proprioceptive cues for motor planning may enhance explicit processing of sensory information, and help retrain sensorimotor loops necessary for recovery of motor function.

Such approaches could be particularly relevant for clinical populations in which proprioceptive deficits contribute to motor impairments, including hemineglect and FMD.

**The potential for tailored movement training following stroke**

As a conceptual exercise, we modeled a precision rehabilitation scenario by calculating the percentage of individuals who achieved the minimal clinically important difference (MCID) in the Box and Blocks Test (≥6 blocks(*48*)) based on training type and baseline proprioceptive status (Fig. 4). In the current standard-of-care scenario (training without considering proprioception) 27% of individuals reach the MCID with Standard training. Virtual Assistance yields a similar 27% response rate, while Propriopixel training increases this to 40%.

In contrast, a precision rehabilitation approach that tailors therapy to proprioceptive status could substantially improve outcomes. Among those with intact proprioception, Standard training achieved a 40% response rate (a 13% gain over the untailored approach), whereas individuals with proprioceptive impairment benefit most from Propriopixel training (63%), followed closely by Virtual Assistance training (42%). These scenarios underscore the potential benefits of tailoring different robotic rehabilitation strategies to individual sensory profiles, supporting the broader vision of precision therapy.

Overall, the results from this study support the hypothesis that proprioceptive impairment is a key predictor of responsiveness to robot-assisted therapy, such that individuals with impaired proprioception are unable to meaningfully benefit from Standard training. Proprioceptively-tailored training, especially Propriopixel training, uniquely improved neural estimation of passive movement sense, likely by refocusing attentional mechanisms to retrain higher-order proprioceptive processing, that improved functional hand outcomes. Further, tailoring training to initial proprioceptive ability may improve training response rates. Specifically, Propriopixel training yielded the greatest benefit for those with proprioceptive deficits, while Standard training was most effective for individuals with intact proprioception. Finally, unlike both physical-assistance-based paradigms, the Virtual Assistance paradigm does not require force generating hardware and can be implemented with simpler, affordable sensor-based systems suitable for remote or home-based rehabilitation, and may provide a generalizable non-robotic training method that can significantly expand access to movement training after stroke.

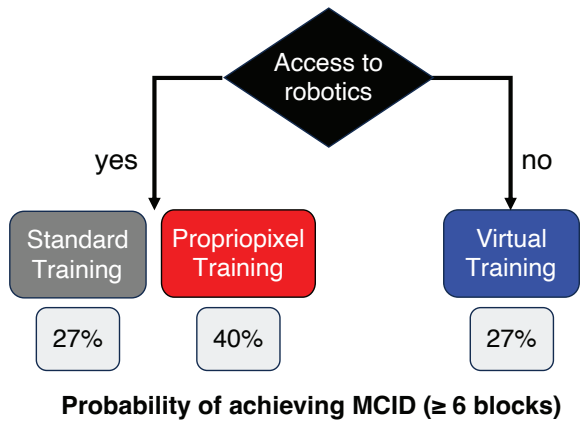
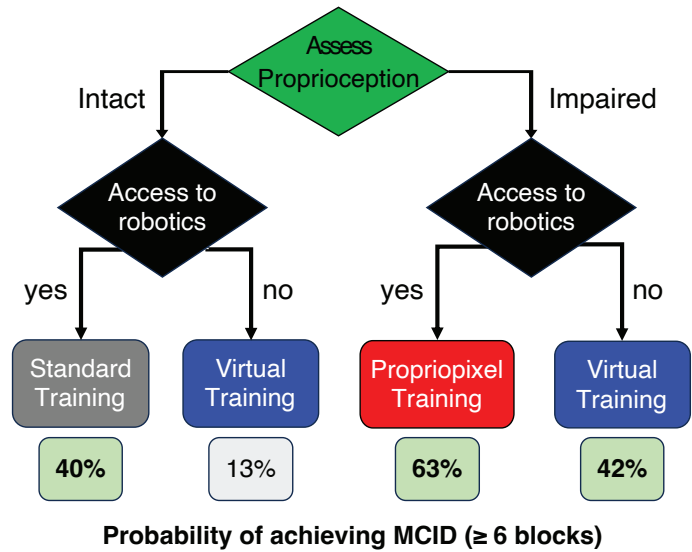

*Figure 4: Flow charts of expected patient response to different forms of training with and without considering proprioception. A: Without Considering Proprioception: Response rate (bubbles under each training type), is calculated as the percentage of participants that reached the minimally clinically important difference (MCID) in BBT (≥6 blocks). In line with the literature, 27-40% of patients achieved clinically meaningful change. B: Considering Proprioception: For individuals without proprioceptive impairments, Standard training yields the highest response rate (40%). For individuals with proprioceptive impairments, Propriopixel training executed with a robot yielded a 63% response rate. Virtual training, executable without a robotic device, yielded a 42% response rate.*

There are some limitations to this study. First, the sample size within groups was relatively small, such that the predictive value of baseline proprioception should be further validated. Additionally, participants with severe hand or cognitive impairments were excluded, highlighting the need to test these approaches in more functionally diverse populations. Consequently, it is unknown whether Virtual Assistance is effective for individuals with greater motor impairment. While the Propriopixel training used physical assistance, it could be adapted to use Virtual Assistance, which may further enhance outcomes. Finally, the generalization of these findings to other limbs, robotic platforms, or sensory systems remains unclear.

# MATERIALS AND METHODS

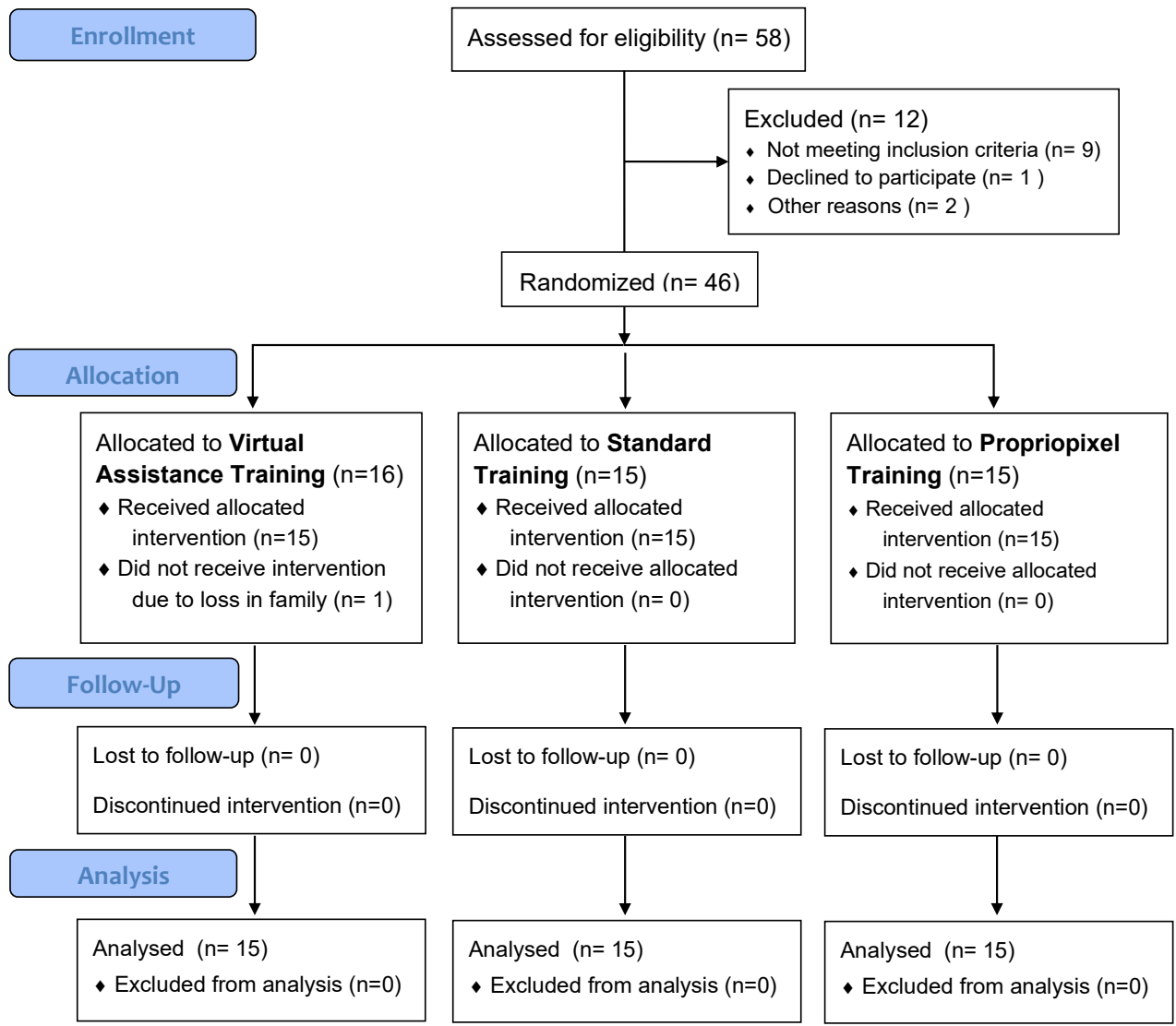

*Figure 5: Flowchart of participant recruitment and completion of training.*

Participants were recruited from the UCI stroke survivor database, regional hospitals, and stroke support groups. Inclusion criteria included age 18-85 years, unilateral UE weakness due to stroke (ischemic or intracerebral hemorrhage, confirmed via radiological imaging) occurring >6 months prior to enrollment. Exclusion criteria included scoring <3 blocks or having <20% difference between limbs on the BBT, severe upper-extremity spasticity, aphasia, or major depression, and concurrent participation in another study related to stroke recovery (see supplemental materials for more details). All visits took place at UCI, and the study was approved by the UCI Institutional Review Board. The trial was registered on ClinicalTrials.Gov (NCT04818073), and all experimental procedures were conducted according to the Declaration of Helsinki. Written and informed consent was obtained.

Prior to the clinical trial, we performed feasibility studies to ensure the suitability of the software and physical interfaces(*42, 68*). Power analysis based on our previous FINGER study(*19*) assumed an effect size of 1.2 on BBT (primary outcome measure) between training groups. Twelve participants in each group (36 total) was expected to provide 80% power at 95% significance-level (two-tailed). We thus recruited 15 per group to allow for participant dropout. One participant dropped out in the first week due to a family emergency (Fig. 5).

The experimental protocol (Fig. 6) included two baseline assessment visits (5-14 days apart), 9 sessions of FINGER robot training across 3 weeks, one post-therapy assessment (1-10 days post training) and a one-month follow-up (1MFU) assessment (25-40 days post training). Participants were block-randomized to a training group (Standard, Propriopixel, or Virtual) by author A.F. at the second baseline visit, based on baseline BBT score and age, using a covariate-adaptive algorithm coded in MATLAB(*19*). The physical therapist performing all clinical assessments was blinded to group assignment. Training sessions with the FINGER robot(*42, 69*) targeted the index, middle finger, and thumb of the affected limb (Fig. 6) and lasted 90-120 mins each. Each session, participants trained in 10 games of RehabHero and 18 games of FingerPong (detailed below) resulting in 1050 flexion/extension movements of the fingers per session. Adverse events were tracked by the treating therapist, and by participant ratings on the visual analog pain scale at the start and end of each session.

The primary clinical outcome measure was change in the BBT score from baseline to 1MFU. BBT is widely used, has good reliability and validity, and quantifies functional dexterity of the UE(*70*). Secondary clinical outcome measures included the Fugl-Meyer Assessment for Upper-Extremity (Motor and Sensory assessments), the 9-Hole Peg Test, the Modified Ashworth Scale of Spasticity, the Motor Activity Log, and the Trail Making Assessment (A and B)(*48, 71*). We additionally quantified performance during unassisted gameplay, finger strength, and finger and thumb proprioception robotically, and measured proprioceptive-related brain activity using EEG (Fig 6C). The BBT, unassisted gameplay and proprioception measures were performed at assessment visits and once per week during training; secondary clinical outcome measures, finger strength and EEG were conducted at assessment visits only (Tbl. S1).

FINGER robot: The FINGER robot facilitates flexion/extension movements of the index and middle fingers, and flexion/extension and adduction/abduction of the thumb(*42*). The training workspace was set to 100% of participants active range of motion in the device and bounded (as applicable) to 90% of their passive range of motion. Consistent with our previous study, all FINGER training games and assessments were performed with an opaque screen covering the hand from view (Fig. 6B-F).

Robotic Rehabilitation games: RehabHero, developed in a previous clinical trial of FINGER(*19*), was modeled after the video game GuitarHero. In RehabHero, participants are cued to hit one of three notes by flexing their either index finger, middle finger, or both to hit a scrolling note as it reaches a target (Fig. 6D). Like GuitarHero, different notes correspond to different finger combinations. Participants played 5 songs with only two notes (top note/index finger; bottom note/middle finger) and 5 songs with all three notes (middle note/both fingers).

Participants also played FingerPong, a new game adapted from the classic Pong computer game (Fig. 6E)(*35*). In FingerPong, participants moved (flexion/extension) their finger to move a paddle (downward/ upwards) to hit a ball back toward a computer-generated opponent (10 games), or to hit computer generated targets (8 games). In each mode, participants played half the games with their index finger, and half with their middle finger.

The Standard training group played both games with a visual interface that displayed all elements of gameplay described above, and received active physical assistance to help them achieve an 80% success rate.

The Virtual training group played both games with the same visual interface as the Standard group, but received "virtual assistance" in which the mapping between robot measured to displayed movements was amplified, and gaming speed and accuracy requirements were adjusted to help participants achieve an 80% success rate.

The Propriopixel training group played both games with a modified visual display, in which some visual gaming cues were replaced with physical cues provided by the robot(*35*, *36*). In RehabHero, visually displayed notes were replaced by a scrolling bar to indicate timing, while the robot moved the thumb to indicate which note (top, middle, or bottom) to hit, requiring players to sense their thumb position to make the correct finger movement (Fig. 6E). Similarly, in FingerPong, the ball was replaced by a scrolling vertical line. The vertical position of the ball (up/down) was displayed by the robot extending(up) or flexing(down) one finger ("ball" finger). Players had to move their other finger that controlled the paddle to hit (match or offset) the "ball finger" (Fig. S1). Thus, participants had to rely on proprioceptive feedback of robot-facilitated movements to play the games, rather than visual information on the screen. Like the Standard group, the Propriopixel group received physical assistance to help achieve 80% success.

Controlling Success Rates: For all assistance strategies, we used a previously tested algorithm(*19*, *41*) to adjust assistance gains to aim for an 80% success rate. Assistance gains were increased by a relative increment of 1 for each unsuccessful movement (i.e., note missed) and decreased by ¼ for each successful movement (i.e., ball hit)(*5*, *41*). If participants exceeded the 80% success rate without assistance (virtual or physical), we incrementally increased the gaming difficulty. Each week, assistance was tuned in the first training session then held constant for the subsequent sessions.

See "Expanded Materials and Methods" in Supplemental Materials for more details.

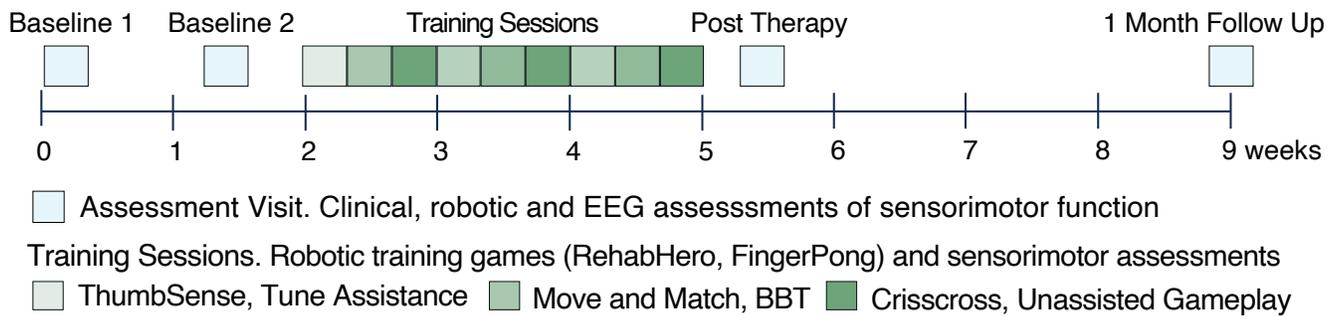
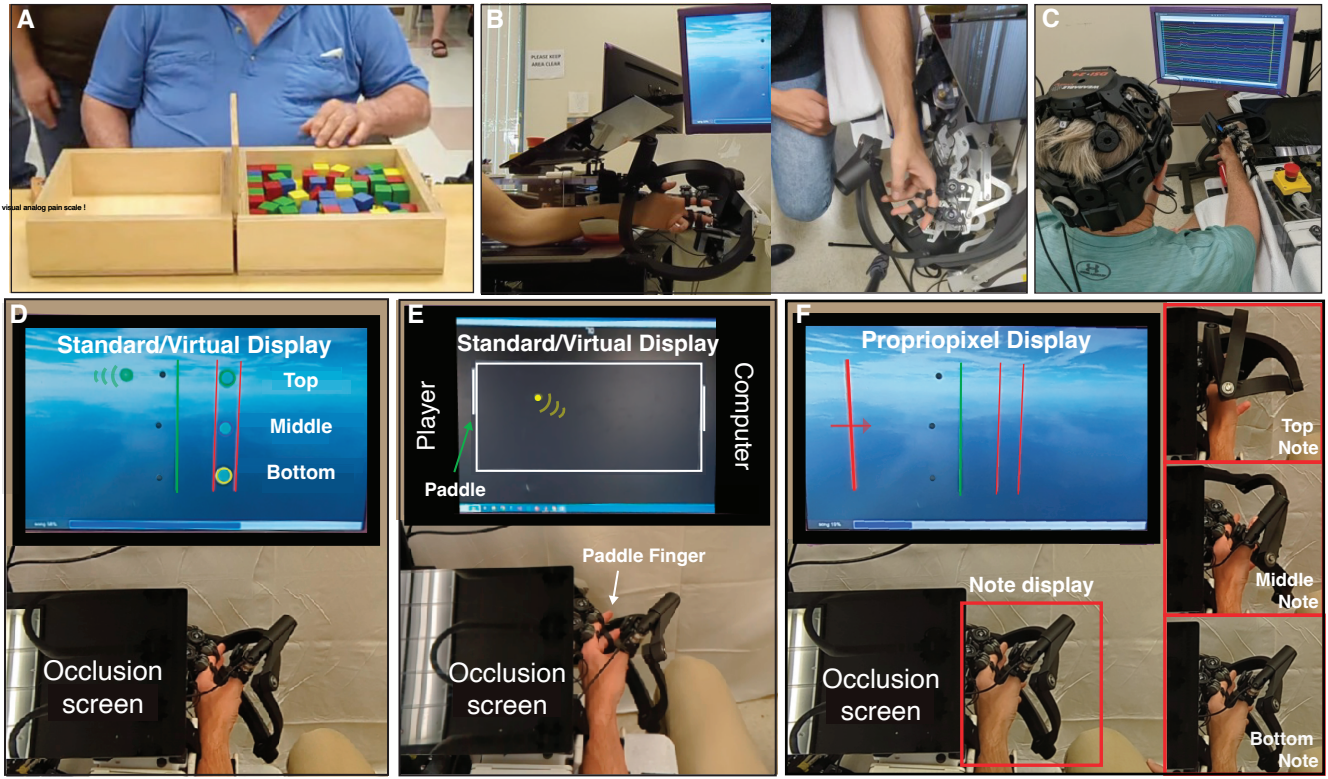

Figure 6: Experimental Protocol. The schedule of longitudinal sensorimotor assessments and assistance tuning during training are detailed in the legend. *A* Box and Blocks Test, the primary outcome, measures the number of blocks transferred over the divider in 60 s. *B.* Hand in the FINGER robot, that measures and assist flexion/extension of the index and middle fingers and thumb circumduction. *C:* Wireless dry EEG system used during Crisscross assessment of finger sensation. *D:* Visually-guided RehabHero. Scrolling notes (green ball) indicate which finger(s) to move to hit the note (top note/index finger, middle note/both fingers, bottom note/middle finger) when it reaches the targets (between the red lines). *E:* Visually guided FingerPong. The paddle (green arrow) and ball (yellow circle) are always visible. The player controls their paddle (down/upwards) to hit the ball via finger flexion/extension. *F:* Propriopixel RehabHero. The robot moves the thumb to indicate the incoming note position (top/middle/bottom) and holds it until the note, now represented by a scrolling red vertical line, reaches the targets.

**Sensorimotor Assessments**

Robotic Assessments of Motor Performance: To quantify changes in finger movement ability, we measured participants' longitudinal success during unassisted gameplay in both rehab games. We additionally measured participants hand capacity(*43*), as the area (force workspace) between the index and middle finger's maximal voluntary contraction, which provides a combined measured of finger strength and individuation.

Robotic Assessments of Proprioception: We performed three assessments of finger proprioception, with vision of the hand occluded. The primary assessment, Crisscross, previously predicted BBT improvement following training with FINGER(*19*). In Crisscross, the robot moves participants' affected index and middle fingers in a crossing pattern between 12-54 degrees of metacarpal phalangeal (MCP) flexion. Participants pressed a button with their unaffected hand when they perceived their fingers crossing. Crisscross had 20 crossings at pseudorandomized speeds between 8-18 deg/s. Performance was quantified as the average absolute error between fingers at button press.

We included two secondary assessments. In Move and Match, participants moved one finger (index/middle) to actively track robot driven movements of the opposite finger (10 deg/s, alternating 12-54 degrees MCP flexion), emulating widely used joint position reproduction assessments(*72*). The required finger tracking movements were similar to Propriopixel FingerPong. Performance was quantified as the average absolute tracking error between fingers. In ThumbSense, FINGER pseudorandomly moved the thumb into one of the three thumb positions used to indicate notes in Propriopixel RehabHero (Fig. 6F). Participants identified each position ("top", "middle" or "bottom") each time the thumb came to rest (6-10s) verbally or via hand signs. Performance was quantified as percent accuracy.

Crisscross Control Group: To establish normative proprioceptive ranges, 37 age-matched controls free from neuromuscular injury (55.6 ± 17.6 years; 19 M; 28 right-handed) performed the same *Crisscross* assessment.

EEG Assessment: To assess neural correlates of proprioceptive processing, EEG was recorded across five consecutive runs of Crisscross (100 total crossings) before (baseline) and after training

(1MFU) using a DSI-24 dry 24-channel headset (Wearable Sensing, CA) and BCI2000 software. Data were recorded via a wired connection at 300Hz and synchronized with robotic data using a common TTL pulse. A 32" monitor (60 Hz refresh rate) placed at eye level ~40 inches from the participant displayed task instructions and feedback. Performance feedback displayed "Difference=X", where X indicated the percent workspace-normalized error between the fingers at button press, shown only after the first press in each crossing. After each crossing, the monitor reset to display: "Press the button at perceived crossing". The time between the monitor reset and the onset of the next crossing was randomized between 1500–3000 ms to ensure the next crossing onset was unpredictable.

Data Analysis: Data were processed in MATLAB 2021a. Robotic data were low-pass (25 Hz) filtered with a fourth-order zero-phase Butterworth filter. Linear actuator data and load cell data were used to calculate MCP joint angles, velocities and torques using the forward kinematics of the FINGER robot(*41*).

EEG data were notch (60 Hz) and bandpass filtered (0.1-30 Hz) using a fourth-order zero-phase shift Butterworth filter, and ICA denoised to remove eye-related artifacts (EEGlab, runICA). Data were then epoched (-250 ms to 3000ms) and baseline corrected (-250ms to -50ms) with respect to movement onset events. Noisy epochs were identified per channel and removed based on trial statistics (see supplemental materials for details), resulting in an average removal of $5 \pm 10\%$ of trials across all channels(*73*). Channels with > 50% removal were excluded from further analysis. Participant's event related potential (ERP) responses were quantified as the average of the remaining epochs.

For participants performing the task with their left (affected) hand, electrode locations were mirrored across the midline. Thus, electrodes displayed over the left hemisphere correspond to the affected hand in the robot, and the right hemisphere corresponds to the hand pushing the button.

ERP magnitude (0.5-1s post movement onset) was analyzed over sensorimotor electrodes (Cz/C3, Pz/P3, and Fz/F3) contralateral to the affected hand. During this time window participants must attend to their finger movement to estimate the finger crossing event, which occured >1s post movement onset for all trials. We examined the change in ERP magnitude between groups and with Crisscross performance.

Statistical Analysis We tested behavioral data (robotic measures, clinical outcomes) for normality with the Anderson-Darling test. For normal data, we used a mixed model analysis with fixed factors assessment timepoint (Baseline, 1MFU) and training mode (Standard, Virtual, Propriopixel), and their interaction, and included random intercepts for each participant to account for initial impairment levels. When significant effects were detected, we used post-hoc Tukey testing to determine factor-level results. For non-normal data, we used repeated measures Friedman's testing to test for the main effect of timepoint on behavior measured across training. We used Kruskal-Wallis testing (non-parametric, one-way ANOVA) to test for the main effect of training group, and to compare change with training between groups (time x training interaction). We performed Wilcoxon testing to determine within-group (signed rank) and between-group (rank sum) effects.

To examine effects of baseline finger proprioceptive impairment on training response (BBT), each training group was divided into subgroups with or without proprioceptive impairment. Proprioceptive Impairment was defined as a Crisscross error more than two standard deviations above the mean of the unimpaired, age-matched controls. In the Standard group, only five individuals met this criterion. To increase statistical power, we included 15 participants from our previous FINGER training trial([19]) who met the same inclusion criteria, completed the same Crisscross assessment, and received nearly identical physically assisted movement training. Procedural differences in training were the addition of passive thumb movement during the RehabHero game and inclusion of the FingerPong game. Both groups completed a comparable number of movements (previous group: 7994 ± 1101, current group: 9079 ± 786). The combined Standard group comprised 12 participants with impaired proprioception and 18 with intact proprioception.

A detailed table regarding the number of participants with available data for each outcome measure is provided in supplemental materials (Tbl. S2).

**List of Supplementary Materials**

Expanded Materials and Methods, Missing Data and Imputation Procedures, Supplemental Results.

Figures S1 to S6

Tables S1 to S4

**Acknowledgments:** Thanks to additional members of the research team: Jessica Bennet, Rubi Tapia, Liesel Blau, Allie Dandel, Anson Chen, and Abraham Ko.

**Funding:**

National Institutes of Health grant R01HD062744 (DJR)

National Institutes of Health grant NCT04818073 (DJR)

**Author contributions:**

Conceptualization: DJR, SCC, EW

Methodology: DJR, SCC, AD, EW, JP, DG, VC, DR, LGF, AJF

Investigation: DJR, VC, JOE, RDR, LGF, AJF

Visualization: DJR, LGF, AJF

Software: AJF, LGF, DR, DG, JP, EW

Funding acquisition: DJR, SCC, EW

Project administration: DJR, SCC, AD, VC

Supervision: DJR, VC, JP, EW, AJF

Writing – original draft: AJF, DJR

Writing – review & editing: AJF, LGF, VC, DG, JP, EW, AD, SCC, DJR

**Competing interests:** Dr. David Reinkensmeyer has financial interest (stock and consulting) in rehabilitation device companies Hocoma and Flint Rehabilitation.

**Data and materials availability:** De-identified data and processing code will be made available upon request to the corresponding author, Dr. Reinkensmeyer.

**SUPPLEMENTAL MATERIALS**

The supplementary materials contain three sections:

1. Expanded Materials and Methods
2. Missing Data and Imputation Procedures
3. Supplemental Results

1. **Expanded Materials and Methods**

Exclusion criteria: Exclusion criteria included scoring <3 blocks or having less than a 20% difference between limbs on the BB(*49*)T, severe upper-extremity spasticity (Modified Ashworth Scale score greater than 3), severe aphasia (NIH Stroke Scale question 9 score 3), evidence of major depression (DSM V criteria or Geriatric Depression Scale score > 10), and concurrent participation in another study related to stroke recovery. Participants also performed the Montreal Cognitive assessment at Baseline (mean 26 ± 3.8, range [14-30]) to screen for cognitive impairment. Individuals with scores below 25, indicative of mild to moderate cognitive impairment, were given a chance to play the games to ensure they were able to follow verbal instructions and play the games. For the participants screened in this study, everyone was able to understand how to play the games with instruction, so no individuals were excluded based on cognitive impairment.

FINGER robot: The FINGER robot can measure and assist flexion/extension movements of the index and middle fingers, and flexion/extension and adduction/abduction of the thumb(*42*). FINGER was controlled via a custom Simulink controller operating at 1000 Hz executed on a Speedgoat performance real-time target computer. The workspace for training was set to 100% of their active range of motion, and restricted (as applicable) to 90% of the participant's maximal passive flexion/extension range of motion in the device. Consistent with our previous study, all FINGER training games and assessments were performed with an opaque plastic screen (occlusion screen, Fig. S1A) covering the hand from view.

Physical Assistance Strategy: Physical assistance was implemented in the same way as the original FINGER study. Briefly, the FINGER robot applied assistive forces with a tunable gain to guide the fingers along a smooth trajectory to intercept the note at the cued time, or to move the paddle to intercept the ball. Assistance was only provided if the participant initiated movement themselves, as determined via load cells mounted behind each finger (threshold = 2N). In RehabHero, assistance gains were finger and direction specific; in FingerPong, assistance gains were averaged across both fingers and movement direction. Physical assistance was provided for the Standard and Propriopixel training group.

Propriopixel Gaming: For the Propriopixel training approach, we replaced some visual gaming cues with physical cues provided by the FINGER robot (*35*, *36*). In RehabHero, we replaced the display of individual notes with a vertical bar moving across the screen to display only note timing but not the string the note was progressing along (see Fig. 6 in main text). The note string (top/middle/bottom) was displayed by having the robot move the thumb (see Fig. 6 in main text, inserts). As with the Standard training group, physical assistance was provided; the robot moved the thumb to radial abduction for top notes, palmar abduction for bottom notes, and halfway between to indicate middle notes. Thus, to play the game, participants had to proprioceptively sense their thumb position to decide which finger(s) to move to hit the incoming note.

Similarly, for FingerPong we replaced the display of the ball's on the screen with a line to indicate the ball's horizontal position on the screen (Fig. S1A). The robot then guided flexion/extension movement of one finger (the "ball finger") to indicate the ball's vertical (bottom/top) position on the screen (Fig S1B). Thus, to play this game, participants had to proprioceptively sense the "ball finger's" (finger controlling the ball) position and then move the "paddle finger" (finger controlling the paddle) to match its position to hit the ball. In the target mode, participants were cued to move the "paddle finger" into relative positions to the "ball finger" (above, matched, below) to hit the ball to computer-generated targets randomly presented at different vertical locations. In both games, players had to attend to facilitated movements to cue self-generated movements of the fingers.

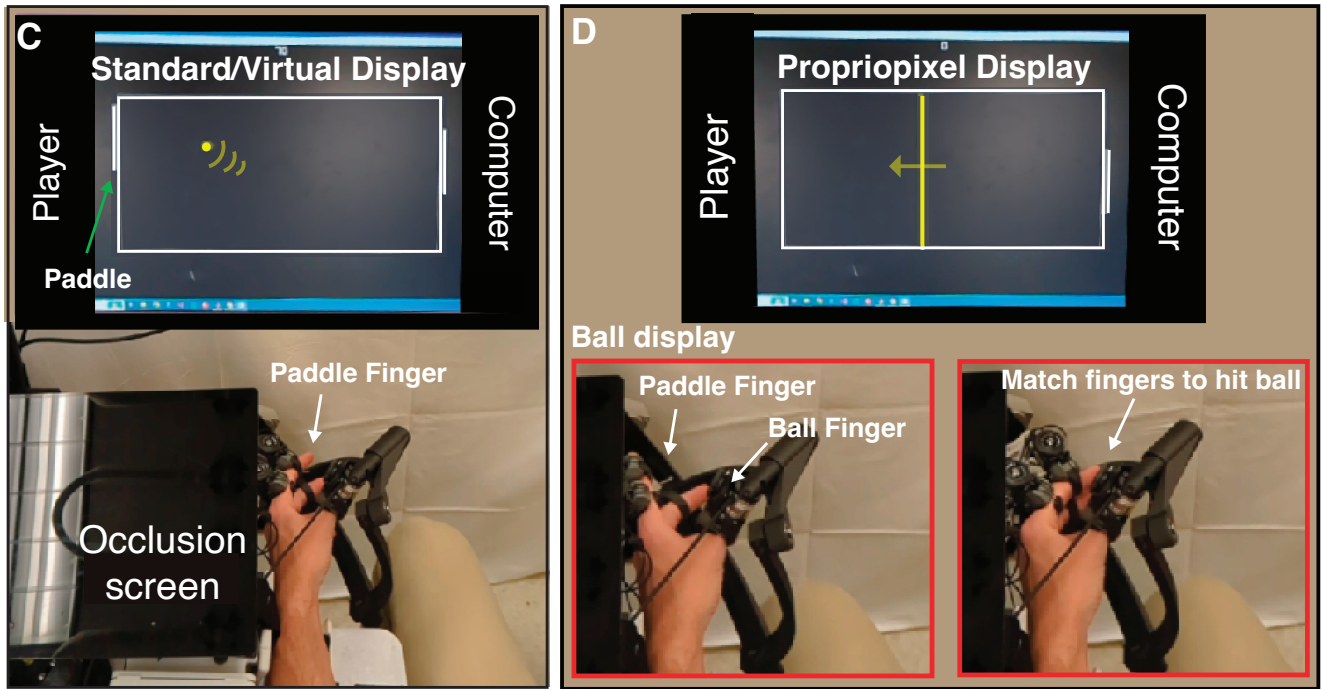

Figure S1: A: Visually guided FingerPong interface as shown in Fig. 6 in the main text. The player's paddle (indicated by the green arrow) and the ball (yellow moving ball in the center of the screen) are visible at all times during gameplay. The player controls their paddle by flexing/extending their finger to move the paddle downward/upward on the gaming monitor. In all training modes, vision of the hand is blocked from view by the black plastic occlusion screen. B: Proprioceptive FingerPong. The ball is replaced by a scrolling line (yellow vertical line) that indicates the ball's horizontal progression across the screen. The vertical position of the ball was displayed by the robot moving one finger, the "ball finger", in flexion/extension that corresponded to the ball's downward/upward position in the game (bottom left insert). The participant had to move their other finger that controlled the paddle , their "paddle finger",  to match their "ball finger" (bottom right insert) when the yellow line reached their side of the monitor in order to hit the ball. For target mode, the interface instructed participants to either match or achieve a relative offset between their paddle and ball fingers to return the ball to specific targets

Virtual Assistance: For the Virtual Assistance training approach, we controlled gameplay success by adjusting the computer-rendered representation of participants' movements and the required timing accuracy during gameplay(*36*). In RehabHero, we assisted players by applying a tunable linear gain that amplified flexion and extension movements made on either side of an inflection point (middle of the workspace), and by adjusting hit timing requirements. Missed notes resulted in both movement and time adjustments, while timing errors resulted in time adjustments only. In FingerPong, virtual assistance was achieved by increasing the paddle size and slowing down the ball. Other than the graphical scaling of finger to game movement and the change in paddle size, the visual interface of the games in physical and virtual assistance modes were identical. In both modes, training was performed with the hand hidden from view, requiring participants to rely on proprioception to map visual representations of movement cues into to accurate motor commands. Physical assistance helped participants perform accurate movements, while virtual assistance adjusted virtual amplification of movement and task success.

Controlling Success Rates: We used the same algorithm tested previously to aim for an 80% success rate for all training modes(*19*, *41*). This was achieved by increasing assistance gains by a relative increment of 1 for each unsuccessful movement (i.e., note missed) and decreasing them by ¼ for each successful movement (i.e., ball hit)(*5*, *41*). If participants exceeded the 80% success rate without assistance (virtual or physical), we incrementally increased the gaming difficulty: in RehabHero, we increased the required note timing accuracy, and in FingerPong we increased the ball speed and decreased the paddle size.

Protocol of Assessments and Robotic Training: A detailed schedule of the clinical and robotic assessments conducted at each timepoint is provided below (Tbl. S1). Motivational assessments were not reported on in this paper, but were examined in our prior manuscript(*36*).

| Assessments | BL1 | BL2 | Weekly Assessments | PT | 1MFU |
|---|---|---|---|---|---|
| **Clinical Assessments** | | | | | |
| NIH Stroke Scale | x | | | | |
| Geriatric Depression Scale | x | | | | |
| MoCa | x | | | | |
| MAL | x | | | x | x |
| Trial Making Test | x | | | x | x |
| Modified Ashwork Spasticity Scale | x | x | | x | x |
| Fugl-Meyer Motor | x | x | | x | x |
| Fugl-Meyer Sensory | x | x | | x | x |
| Nine Hole Peg Test | x | x | | x | x |
| Visual Analoge Pain Scale | x | x | every session | x | x |
| Box and Blocks | x | x | sessions 2,5,8 | x | x |
| **Robotic Assessments** | | | | | |
| Maximum Voluntary Contraction | x | | | x | x |
| Reaction Time | | x | sessions 1,4,7 | x | x |
| GuitarHero Unassisted | | x | sessions 3,6,9 | x | x |
| FingerPong Unassisted | | x | sessions 3,6,9 | x | x |
| GuitarHero Tune Assistance | | | sessions 1,4,7 | | |
| FingerPong Tune Assistance | | | sessions 1,4,7 | | |
| ThumbSense | x | x | sessions 1,4,7 | x | x |
| Move and Match | x | x | sessions 2,5,8 | x | x |
| Crisscross | x | x | sessions 3,6,9 | x | x |
| **EEG Assessments** | | | | | |
| Resting State Connectivity | | x | | x | x |
| Sensory Evoked Potential | | x | | x | x |
| Crisscross with feedback | | x | | x | x |
| **Motivational Assessments** | | | | | |
| BBT Self Efficacy | | x | sessions 2,5,8 | x | x |
| GuitarHero Self Efficacy | | x | sessions 3,6,9 | x | x |
| FingerPong SelfEfficacy | | x | sessions 3,6,9 | x | x |
| Intrinsic Motivation Inventory | | x | sessions 1,4,7 | x | x |

Table S2: Table detailing what sensorimotor assessments were performed at each experimental time point. Weekly assessment column details the specific session order that longitudinal assessments were performed on.

EEG Assessment and Data Processing:

To investigate the neural correlates of proprioceptive processing, we collected EEG data across five consecutive runs of Crisscross (100 total crossings) before (baseline) and after therapy (one-month follow-up). We measured EEG using a DSI-24 dry 24-channel headset (Wearable Sensing, CA) and BCI2000 acquisition software. Data were recorded at 300 Hz via a wired connection and synchronized with robotic data using a common TTL pulse. A 32" monitor with a 60 Hz refresh rate was placed at eye level ~40 inches from the participant to display task instructions and feedback. Performance feedback displayed "Difference = X", where X was the percent error magnitude (angular error between the fingers at button press, normalized by the task workspace), and was provided only after the initial button press per crossing. Due to the sampling frequency of the controller and screen refresh rate of the monitor, the latency between button press and feedback displayed on the monitor was roughly 83 ms. At the end of each crossing, the monitor reset to display the instructions: "Press the button at perceived crossing". The time between the monitor reset and the onset of the next crossing movement was randomized between 1500–3000 ms to ensure that the timing and onset of the crossing was unpredictable.

EEG data were notch filtered at 60 Hz and then bandpass filtered between 0.1 and 30 Hz using a fourth-order zero-phase shift Butterworth filter. Independent component analysis (ICA) with 19 components was performed using the runica algorithm in EEGlab. Component time course and topographical maps were visually inspected to identify and remove eye-related artifacts. Data were then epoched (-250 ms to 3000ms) and baseline corrected (-250ms to -50ms) with respect to movement onset events. All epochs with an absolute voltage exceeding 100 mV, or a timepoint to timepoint change in magnitude of 50 mV were excluded. Finally, noisy epochs were identified per channel and removed based on trial statistics: if the median voltage or variance of the epoch was greater than 3 median absolute deviations from the median voltage and variance measured across all epochs for that channel (*73*). This resulted in an average removal of 5 $\pm$ 10% of trials across all channels, and an average of 4.2 $\pm$ 8% trials (range [0-50%]) for the sensorimotor electrodes of interest (F3,C3,P3,Cz,Fz,Pz).

Channels with less than 50% remaining trials were excluded from further analysis (1 Fz electrode for one participant, and 1 Cz electrode for a different participant). Each participant's event related potential (ERP) response was quantified as the average of the remaining epochs. To assess responses related to proprioceptive estimation, we measured the ERP magnitude within a 0.5 to +1 second window after movement onset, prior to any finger-crossing events, in which participants must attend to finger movement to estimate the finger crossing event. Measurements were taken from sensorimotor electrodes (Cz/C3, Pz/P3, and Fz/F3) over the brain areas associated with the hand performing the proprioceptive task. We investigated the change in ERP magnitude between groups, and how the magnitude of this signal varied with Crisscross task performance.

## 2. Missing Data and Imputation Procedures

|  | Missing/Excluded Subject Data | | |
|---|---|---|---|
| Assessment | Baseline | Post Therapy Assessment | 1 Month Follow Up |
| Gameplay Assessment | 1,6,18,33,35,40 | - | - |
| Hand Capacity | 16,17,26,31,37 | 26,31,37,39 | 11,15,26,37 |
| Crisscross | 16, **41** | 20,**41** | **41** |
| ThumbSense | 16,17 | - | - |
| Move and Match | 1,6,16,18,20,24,33,35,37,39,40 | 1,6,18,24,33,35,37,39,40 | 1,6,18,24,33,35,37,39,40 |
| EEG Analysis | 16,17,25,31,**33**,41 | 2,11,20,28,31,36,41,**45** | 10,31,36,**40**,41 |
| Virtual Participants: 1,11,14,17,21,23,27,30,35,36,38,42,43,45,46 | | | |
| Propriopixel Participants: 3,5,6,10,12,13,15,16,19,26,28,33,34,37,39 | | | |
| Standard Participants: 2,4,8,9,18,20,22,24,25,29,31,32,40,41,44 | | | |

Table S2: Subject ID numbers of the participants with missing or excluded data for each robotic and EEG assessment of sensorimotor function. Participants whose data was excluded are bolded.

Above is a table of participant data excluded or missing from analysis (Tbl. S3). Subject ID numbers were randomized, and do not correspond to chronological order of data acquisition. The respective group assignments are listed below and color coded for ease of visualizing excluded or missing data within each group. There were no missing or excluded data for any of the clinical assessments.

For robotic assessments, missing data were due to failed data storage by the computer, hardware failure operating FINGER, or the participants' inability to complete a session (e.g. time constraints, fatigue, unscheduled fire alarm) resulting in missed assessments. Move and Match was added to the protocol after running the first 7 subjects, contributing to the high amount of missing data for that assessment. Similarly, Baseline gameplay assessments were added after the first 6 participants, which were initially conducted from week 1 of training onwards. For crisscross, 1 participant (41) was excluded due to an inability to understand the assessment.

For individuals with missing baseline data, we estimated baseline performance in Gameplay, Crisscross and Move and Match using week 1 data for analyses of change with training. This was a conservative estimate, as participant tended to improve in week one compared to baseline, and early change was comparable across training groups. However, these participants were excluded from correlational analysis involving baseline performance, as the exact baseline performance was unknown.

For EEG data, missing data represents participants who either were excluded due to an inability to do the crisscross assessment (41) or had excessive artifacts during acquisition (2, 28, 31, 33, 40, 45), or who experienced hardware failures that prevented data acquisition.

## 3. Supplemental Results

A table (Tbl. S3) reporting the expanded statistical results, including all longitudinal timepoints, as well as differences in training response at the post training time point, is reported below. A figure (Fig. S2) depicting longitudinal change from baseline for our primary outcome measure (BBT), the commonly used Fugl-Meyer Assessment of Upper Extremity Motor and Sensory ability, and all sensorimotor robotic assessments is also provided. Figures report mean and standard error of the mean for visualization purposes, while the table reports median and IQR summary statistics, as the data were non-normally distributed. Effects tended to be larger at the immediate end of training timepoint, however all results were in the same direction as those observed at the primary one month follow up time point.

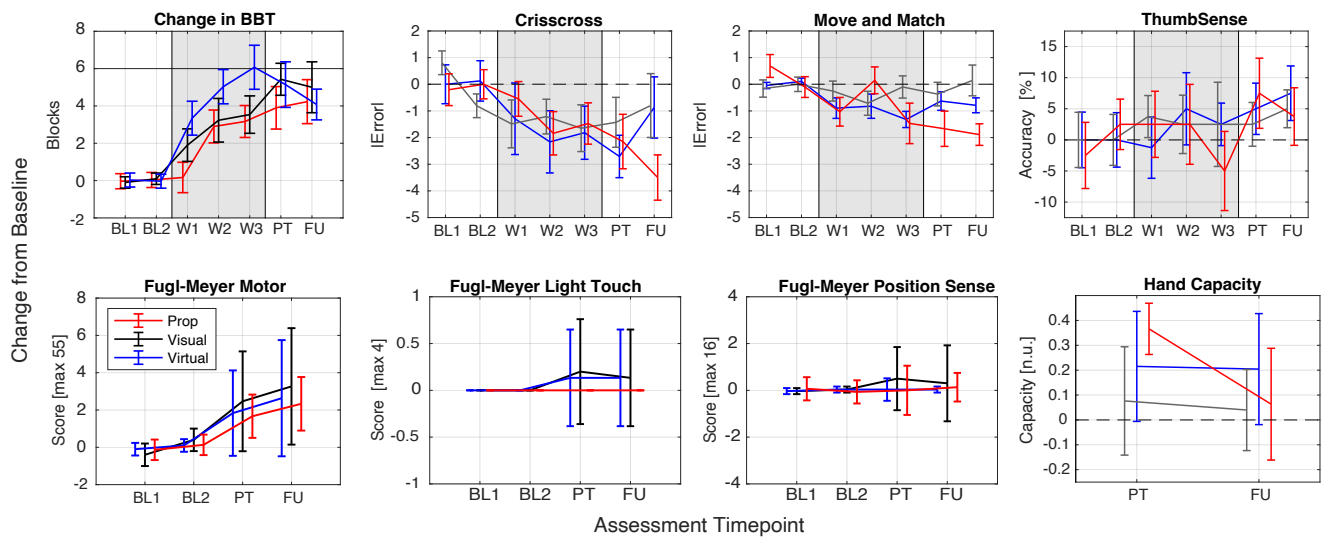

Figure S2: Mean and standard error of longitudinal change with training from baseline for primary clinical (Box and blocks test (BBT), and Fugl-Meyer Assessment of the Upper Extremity) and robotic Sensorimotor assessments (Crisscross, Move and Match, ThumbSense, Hand Capacity). All results represent change from average baseline

performance, except for Hand Capacity, which only had a single baseline assessment, and therefore shows change from baseline (0).

| | Fixed Effects | | Post Training | | | | 1 Month Follow Up | | | |
|---|---|---|---|---|---|---|---|---|---|---|
| | Timepoint | Group | Standard | Virtual | Propriopixel | Interaction | Standard | Virtual | Propriopixel | Interaction |
| **Clinical Assessments** | χ2(DF), α | | median change [range] | | | χ2(DF), α | median change [range] | | | χ2(DF), α |
| Box and blocks | **59.8(5), p<0.0001** | 0.91(2), p<0.635 | 5.50 [0.50,11.50] | 3.50 [-1.00,18.00] | 4.00 [-1.00,12.00] | 0.58(2), p<0.749 | 1.50 [-1.00,15.50] | 3.00 [-2.00,11.00] | 3.00 [-2.50,14.00] | 1.52(2), p<0.469 |
| UE Fugl-Meyer Motor (/66) | **52.7(2), p<0.0001** | 5.45(2), p<0.065 | 2.00. [-1.00,10.50] | 1.00 [-0.50,6.50] | 1.50 [-0.50,3.50] | 1.14(2), p<0.565 | 4.00 [-2.50,10.50] | 2.00 [-0.50,9.50] | 2.50 [0.00,5.00] | 1.38(2), p<0.501 |
| UE Fugl-Meyer Light touch (/4) | 4.67(2), p<0.097 | 0.59(2), p<0.74 | 0.00 [0.00,2.00] | 0.00 [0.00,2.00] | 0.00 [0.00,0.00] | 2.05(2), p<0.359 | 0.00 [0.00,2.00] | 0.00 [0.00,2.00] | 0.00 [0.00,0.00] | 1.02(2), p<0.599 |
| UE Fugl-Meyer Position sense (/16) | 1.00(2), p<0.606 | 0.83(2), p<0.66 | 0.00 [0.00,5.00] | 0.00 [-1.00,1.50] | 0.00 [-2.00,2.00] | 1.67(2), p<0.43 | 0.00 [-1.50,6.00] | 0.00 [0.00,0.50] | 0.00 [-1.00,1.50] | 0.23(1), p<0.89 |
| Motor Activity Log (how much, /5) | **16.5(2), p<0.0003** | 4.3(2), p<0.116 | 0.57 [-0.42,2.17] | 0.67 [-0.23,3.38] | 0.17 [-1.12,1.63] | 5.92(2), p<0.052 | 0.74 [-0.63,3.13] | 0.87 [-0.46,3.23] | 0.19 [-1.25,2.38] | 3.32(2), p<0.19 |
| Motor Activity Log (how well, /5) | **26.6(2), p<0.0001** | 5.66(2), p<0.059 | 0.50 [-0.40,2.38] | 0.62 [-0.25,3.67] | 0.32 [-0.93,1.59] | 4.98(2), p<0.083 | 0.68 [-0.36,3.32] | 0.76 [-0.20,3.15] | 0.16 [-0.53,2.27] | 3.12(2), p<0.21 |
| Nine Hole Peg [pegs/min] | 3.00(2), p<0.223 | 0.21(2), p<0.89 | 0.00 [-10.97,0.39] | 0.00 [-4.77,4.05] | 0.00 [-13.44,3.86] | 3.23(2), p<0.199 | 0.00 [-11.24,2.73] | 0.00 [-9.45,2.95] | 0.00 [-12.09,1.93] | 0.11(2), p<0.946 |
| Trail Making A [s] | **14.9(2), p<0.0006** | 0.65(2), p<0.72 | **-2.66 [-36.22,52.06]** | -3.73 [-15.69,77.90] | -1.00 [-15.81,36.78] | 0.50(2), p>0.59 | -4.44 [-18.74,64.93] | -6.88 [-28.97,96.59] | -6.00 [-20.72,34.62] | 1.03(2), p<0.598 |
| Trail Making B [s] | **6.93(2), p<0.0312** | 0.02(2), p<0.992 | **-16.41 [-93.2,61.1]** | -0.44 [-61.4,101.94] | 11.13 [-21.65,85.7] | **6.15(2), p<0.046** | -14.1 [-37.8,81.85] | -7.26 [-66.7,203.50] | **-6.88 [-106.2,31.3]** | 0.03(2), p<0.985 |
| MASS All Flexors | 4.24(2), p<0.12 | 5.65(2), p<0.059 | 0.00 [-0.40,0.40] | 0.00 [-0.30,0.20] | 0.00 [-0.60,0.40] | 1.71(2), p<0.43 | 0.00 [-0.60,0.40] | 0.00 [-0.30,0.00] | 0.00 [-0.40,0.40] | 1.56(2), p<0.458 |
| MASS All Extensors | 0.00(2), p=1 | 0.00(2), p=1 | 0.0 [0.0, 0.0] | 0.0 [0.0, 0.0] | 0.0 [0.0, 0.0] | 0.00(2), p=1 | 0.0 [0.0, 0.0] | 0.0 [0.0, 0.0] | 0.0 [0.0, 0.0] | 0.00(2), p=1 |
| Visual Analoge Pain Scale (/10) | 3.76(5), p<0.58 | 0.22(2), p<0.893 | 0.00 [0.00,1.50] | 0.00 [-2.25,0.00] | 0.00 [-2.25,0.00] | 3.55(2), p<0.17 | 0.00 [-1.00,0.00] | 0.00 [-1.00,1.25] | 0.00 [-2.25,0.00] | 0.00(2), p<0.99 |
| **Robotic Assessments** | χ2(DF), α | | median change [range] | | | χ2(DF), α | median change [range] | | | χ2(DF), α |
| Hand capacity [n.u.] | **10.46(2), p<0.0054** | 2.67(2), p<0.26 | 0.08 [-1.62,2.14] | **0.22 [-1.25,2.24]** | 0.37 [-0.04,1.12] | 1.13(2), p<0.569 | 0.04 [-0.38,1.89] | 0.20 [-0.42,2.20] | 0.06 [-1.13,2.05] | 0.18(2), p < 0.912 |
| Crisscross [deg] | **23.47(5), p<0.0003** | 1.43(2), p<0.49 | -1.43 [-9.66,5.36] | **-2.71 [-10.72,2.09]** | -2.15 [-6.92,7.53] | 0.92(2), p<0.628 | -0.80 [-17.14,2.81] | **-0.87 [-11.30,7.90]** | -3.50 [-5.99,4.65] | 1.30(2), p<0.522 |
| Move and Match [deg] | **16.21(5), p<0.0063** | 0.04(2), p<0.98 | -0.37 [-2.66,1.83] | -0.63 [-2.03,1.90] | **-1.67 [-7.51,-0.24]** | 5.10(2), p<0.079 | 0.14 [-2.34,4.70] | -0.79 [-2.20,1.38] | **-1.89 [-4.52,-0.25]** | **11.71(2), p<0.0029** |
| ThumbSense [% accurate] | **15.33(5), p<0.009** | 1.41(2), p<0.494 | 0.50 [-5.50,4.00] | **1.00 [-4.50,5.50]** | 1.50 [-5.00,8.50] | 1.36(2), p<0.51 | 1.00 [-4.50,7.50] | **1.50 [-6.00,5.50]** | 0.75 [-3.50,6.00] | 0.51(2), p<0.78 |

Table S3: Extended statistical analysis for all clinical and robotic measures of sensorimotor hand function. Fixed effects of timepoint reports the main effect of timepoint across all groups, using Friedmans repeated measures testing. The main effect of Group was determined via Kruskall-Wallis testing (non-parametric, one-way ANOVA) of the average performance measured across all timepoints. To compare change with training between groups (time x training interaction) we quantified the change from baseline behavior at both the post training and 1 month Follow up (1MFu), and performed Kruskall-Wallis testing to test for differences between training groups at each timepoint. We performed Wilcoxon signed rank testing to determine within group effects and Wilcoxon rank sum testing to determine between group effects. Effects that reached significance at the p<0.05 level are bolded.

**Ceiling effects in the Fugl-Meyer Assessment of Upper Extremity Sensation.**

The FM-Sensory score, a coarse clinical scale, showed no significant change at the group level, but this may have been due to ceiling effects. The Fugl-Meyer Assessment of Upper Extremity Sensation

(FMA-UES) evaluates the ability of participants to detect the presence (yes/no) of light touch to either the finger or hand with a cotton ball (Max score 4), or the movement direction (up/down) of each of the fingers (Max score 16). This assessment lends itself to measurement error due to guessing, particularly for movement direction detection.

For participants in our cohort, at baseline 30 had a perfect (20/20) FMA-UES score, 6 had FMA-UES scores of 18 or 19, and 9 had scores of 4-13, indicative of moderate to severe proprioceptive deficits. These high baseline scores left limited to no room for improvement (ceiling effects), which may have contributed to that lack of significant group level improvement in FMA-UES scores with training. Of the individuals with some level of baseline FMA-UES impairment (FMA-UES < 20), 7 participants improved their scores at 1MFU with respect to baseline (mean $\pm$ std: 2 $\pm$1.8), 5 had no change, and 3 got worse (1 $\pm$ 0.5). However, only 4 of these individuals exhibited changes at 1MFU from baseline that were larger than the changes they exhibited between the baseline 1 and baseline 2 assessments, bringing into question the sensitivity of the FMA-UES in detecting change with training beyond the variability occurring naturally between repeated assessments in the absence of training (Tbl. S4).

All participants with a less than perfect FMA score had impaired proprioception as assessed by Crisscross in at least one baseline assessment. Crisscross identified an additional 5 participants with proprioceptive impairment. Proprioceptive impairment in Crisscross was determined as performance errors exceeding 2 standard deviations above the average of performance errors measured in an aged matched, unaffected control group (N=37). At the group level, there were significant correlations between FMA-UES and Crisscross scores ($r > 0.4$, $p < 0.008$ for all timepoints). However, within the 15 individuals with FMA-UES impairment (Tbl. S4) there was a poor correspondence between FMA-UES scores and Crisscross, suggesting evaluation of these individuals may be more susceptible to measurement errors due to inconsistent estimation (possibly guessing) leading to greater variability between assessments.

In sum, our results suggest that the FMA-UES is less sensitive to detecting proprioceptive deficit at baseline compared to Crisscross, and consequently in detecting change with training due to ceiling effects and high variability in the assessment.

| Crisscross Errors [deg] | | Fugl-Meyer Assessment of Upper Extremity Sensation [score, max 20] | | | | |
| --- | --- | --- | --- | --- | --- | --- |
| Baseline 1 | Baseline 2 | Baseline 1 | Baseline 2 | 1MFU | Baseline 2-1 | 1MFU- Baseline |
| NaN | 16.91 | 18 | 20 | 20 | 2 | 0 |
| 19.20 | 15.04 | 20 | 18 | 17 | -2 | -3 |
| 14.23 | 19.49 | 11 | 12 | 13 | 1 | 1 |
| 12.83 | 19.79 | 10 | 11 | 12 | 1 | 1 |
| 23.51 | NaN | 11 | 11 | 10 | 0 | -1 |
| 22.51 | 16.21 | 9 | 9 | 9 | 0 | 0 |
| 16.39 | 20.92 | 4 | 4 | 6 | **0** | **2** |
| 22.65 | 18.01 | 7 | 8 | 8 | 1 | 0 |
| 21.30 | 21.16 | 18 | 18 | 20 | **0** | **2** |
| 11.50 | 13.69 | 19 | 17 | 20 | -2 | 1 |
| NaN | 23.43 | 19 | 19 | 19 | 0 | 0 |
| 14.83 | 19.47 | 14 | 14 | 14 | 0 | 0 |
| 17.99 | 20.80 | 13 | 12 | 12 | -1 | -1 |
| 17.89 | 27.32 | 18 | 18 | 20 | **0** | **2** |
| 13.31 | 11.69 | 12 | 12 | 18 | **0** | **6** |

Table S4. Comparison of Crisscross Assessment performance (average absolute error [deg]) and Fugl-Meyer Assessment of Upper Extremity Sensation (FMA-UES) scores at both baseline time points, and the measured change in FMA-UES with training. FMA-UES were less sensitive to deficits compared to Crisscross, with little room for detected change with training. Variability between baseline sessions was largely equivalent with the change observed with training.

**Relationship of baseline crisscross to change in BBT**

As reported in the main text, while individuals with poor crisscross performance had poor response (change in BBT) to standard training, Propriopixel training showed the opposite trend, and virtual training showed no relationship to baseline proprioception (Fig. S3).

To complement the correlational analysis we subdivided each training group into individuals with or without baseline finger proprioceptive impairment based on their baseline Crisscross error. Baseline proprioceptive impairment was shown to predict response to different training modes (main text, Fig. 2). To ensure these changes did not reflect differences in baseline hand function consistent with known patterns of proportional recovery, we compared baseline BBT score between impaired and unimpaired training sub-groups.

For Standard and Propriopixel training groups (which exhibited differential training response based on proprioceptive impairment), there was no significant difference between baseline hand function (BBT) between impaired and unimpaired individuals (Wilcoxon ranksum: Standard: z=-1.18, p=0.24, Propriopixel: z=0, p=0.98). Moreover, there were no significant differences in baseline BBT between training groups for individuals with impaired proprioception (Kruskal-Wallis, $\chi^2(2)=1.03$, p=0.60) nor intact proprioception ($\chi^2(2)=0.78$, p=0.68). Thus, the observed differential training response between groups was not due to differential baseline hand function. These results were similarly observed quantifying hand motor function using Fugl-Meyer Motor scores (p>0.29).

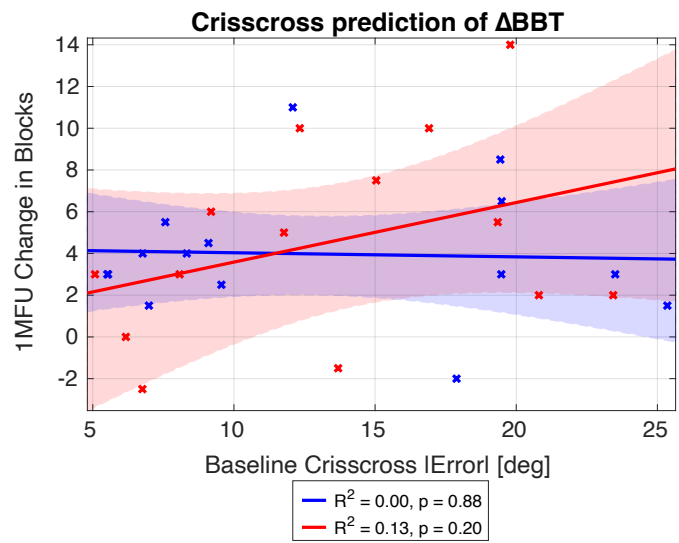

Figure S3: Linear regression between baseline CC performance and change in blocks transferred in the box and blocks assessments, the primary outcome measure.

**Proprioceptive CNV:**

We sought to gain insight into the neural correlates of improved finger proprioception observed in the Crisscross assessment following robotic training. The EEG crisscross assessment included performance feedback to increase task engagement. Behavioral results showed the Propriopixel and Virtual assistance groups significantly improved task performance at 1MFU, while the Standard group did not (Fig. S4).

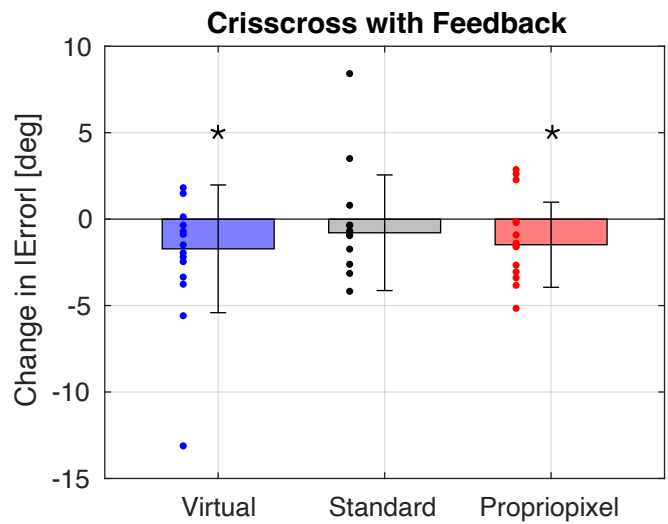

Figure S4: Change in performance across training (1MFU-baseline performance) in the EEG Crisscross task, which included performance feedback. The bar plot depicts the group median response, and errorbars report the group-level standard deviation. Dots to the left of the errorbars report individual participant data.

Our analysis off EEG focused on the Contingent Negative Variation (CNV), an event-related potential characterized by a slow negative wave that occurs when an action (in this case, the button press at perceived finger crossing) is contingent on a preceding stimulus (in this case, proprioceptively sensed movement onset) and reflects the participant's anticipation of the upcoming action (*44, 45*).

EEG recordings during Crisscross revealed a pCNV response: a gradual negative shift in voltage over the sensorimotor electrodes of the hand experiencing the proprioceptive test, time-locked to movement onset and continuing until the button press, after which the signal rebounded positively (Fig. S5). The magnitude of the pCNV response was observed to follow movement progression of the fingers, with the apex of the pCNV occurring over parietal areas (P3, Pz), suggesting that these signals may originate in the sensory cortex. The pCNV was observed to be more gradual with slow crossings, compared to fast crossings, as shown in Fig. S5, A-C.

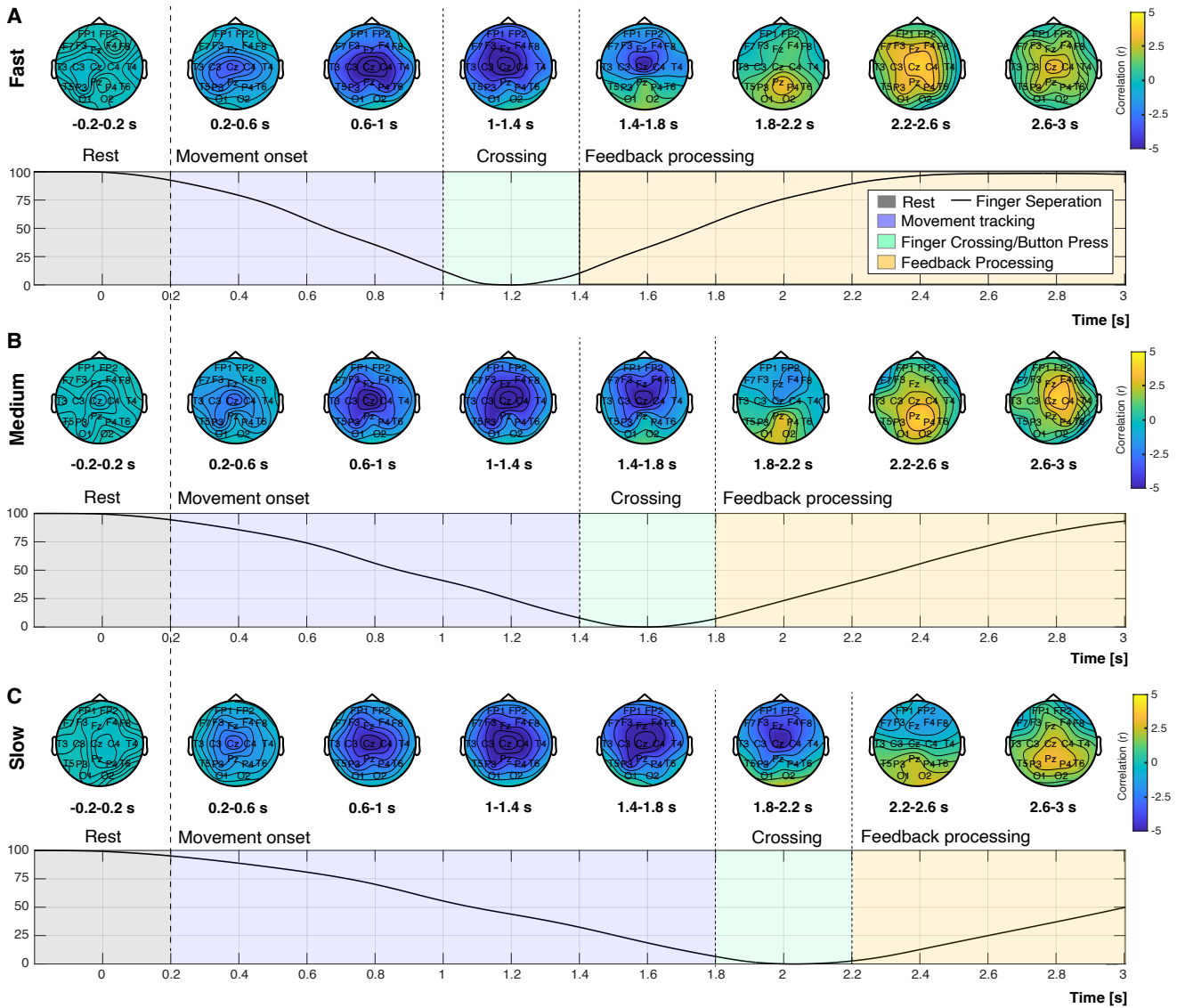

Figure S5: A proprioceptive Contingent Negative Variation (pCNV), observed to progress with finger crossings of different speeds. The average pCNV is shown for individuals with intact proprioception for fast (*A*), medium (*B*), and slow (*C*) speed trials for the Crisscross assessment. In each case, the negative deflection began just after the robot began moving the passive fingers (time = 0) and became increasingly negative before reaching maximum negative deflection around the moment the participant pushed the button, which roughly corresponded to the actual moment of finger crossing. After this, the pCNV rebounded positively consistent with EEG patterns associated with feedback processing. For participants performing the task with their left hand, we flipped the electrodes between the left and the right hemisphere, such that all data had a "right hand configuration" with respect to task execution. This transformation made it so that electrodes over the left hemisphere corresponded to the hand in the robot, while the right hemisphere corresponded to the hand pushing the button.

## Neural Association to Crisscross Task Measures

Given that the pCNV response was observed to vary with trials of different speeds, we conducted a supplemental correlational analysis between task parameters (e.g., finger position, velocity, and positional error) and the recorded EEG time series measured across the entire task (Fig. S6) to understand which Crisscross task features explained neural response. Among the tested variables, the **relative finger velocity** (combined velocity of fingers 1 and 2) prior to the button press showed the strongest association with EEG activity measured within the same topographical regions as the pCNV (Fig. S6, "Relative Velocity"). Finger separation (difference in angular position between finger 1 and 2), was also strongly associated with EEG signals measured over the sensorimotor areas. In our cohort, relative velocity produced slightly larger (2-5%) and more consistent (44-50% fewer outliers) significant associations to neural activity measured across participants.

Crisscross requires participants to estimate the impending crossing and pre-emptively send the motor command to push the button such that the action coincides with the actual crossing event. Accurate task performance therefore depends on precise estimation of movement velocity. Thus, improvements in Crisscross were likely due to training improvement of velocity estimation, evidence by the increased association to velocity following training, especially for the proprioceptive-focus training modes.

It is worth noting that in our task, we cannot definitively distinguish between velocity and position-based contributions to proprioceptive estimation of movement progression. Finger separation was strongly correlated with relative velocity (mean ± std across subjects: r=0.79±0.02), and their collinearity limits our ability to distinguish effects associated independently with velocity or positional information. Moreover, prior studies have shown that the CNS uses both positional and velocity information for proprioceptive estimates of passive limb movements. Therefore, we cannot rule out that positional estimation may also have improved with training and contributed to improved Crisscross performance. Future studies should modify the crisscross task design to better differentiate between independent contributions of velocity and positional signals to neural estimates of movement (e.g. variable finger displacement between trials with constant velocity and vis versa), to determine how training modifies gains in either or both estimation processes.

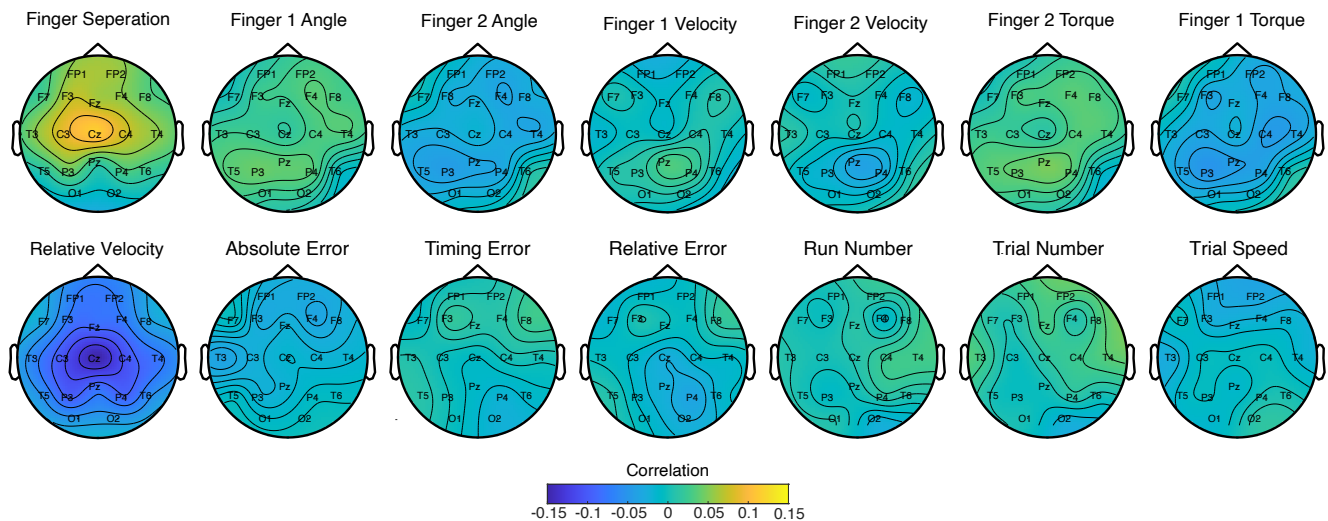

Figure S6: Topography of the association (r) between neural signal measured during the crisscross task and Crisscross task kinematics, kinetics, and performance parameters. Only finger separation and relative velocity showed significant associations to neural activity. These regions overlapped with the topography of the pCNV response.

**Supplement References (order included in the main text)**

1. *(49).* Liang, Kai Jie Chen, Hao LingShieh, Jeng Yi& Wang, Tien Ni. Measurement properties of the box and block test in children with unilateral cerebral palsy. Sci. Rep. 11, 1–8 (2021).

2. *(42).* Ketkar, Vishwanath D. Wolbrecht, Eric T. Perry, Joel C.& Farrens, Andria. Design and Development of a Spherical 5-Bar Thumb Exoskeleton Mechanism for Poststroke Rehabilitation. J. Med. Device. 17, (2023).

3. *(35).* Reinsdorf, Dylan S. Mahan, Erin E.& Reinkensmeyer, David J. Proprioceptive Gaming: Making Finger Sensation Training Intense and Engaging with the P-Pong Game and PINKIE Robot. Annu. Int. Conf. IEEE Eng. Med. Biol. Soc. IEEE Eng. Med. Biol. Soc. Annu. Int. Conf. 2021, 6715–6720 (2021).